\documentclass[lettersize,journal]{IEEEtran}
\usepackage{amsmath,amsfonts}
\usepackage{algorithm}
\usepackage{array}
\usepackage{textcomp}
\usepackage{stfloats}
\usepackage{url}
\usepackage{verbatim}
\usepackage{graphicx}   %插入图片的宏包 
\usepackage{cite}
\usepackage{algpseudocode}
\usepackage{diagbox}
\usepackage{color}
\usepackage{pdfpages}
\usepackage{multirow}
\usepackage{amssymb} 
\usepackage{soul}
\usepackage{tabularray}
\usepackage{subcaption}
\usepackage[colorlinks, linkcolor=red, anchorcolor=blue, citecolor=green]{hyperref}
\renewcommand{\algorithmicrequire}{\textbf{Input:}}
\renewcommand{\algorithmicensure}{\textbf{Output:}}
\begin{document}

\title{MUS-CDB: Mixed Uncertainty Sampling with Class Distribution Balancing for Active Annotation in Aerial Object Detection}
\author{Dong~Liang,
         Jing-Wei~Zhang,
         Ying-Peng Tang,
         Sheng-Jun Huang
\thanks{All authors are with the College of Computer Science and Technology, Nanjing University of Aeronautics and Astronautics, MIIT Key Laboratory of Pattern Analysis and Machine Intelligence, Collaborative Innovation Center of Novel Software Technology and Industrialization, Nanjing 211106, China.  }% <-this % stops a space

\thanks{Manuscript received December 3, 2022.}}

% The paper headers
% \markboth{Journal of \LaTeX\ Class Files,~Vol.~14, No.~8, December~2022}%
% {Shell \MakeLowercase{\textit{et al.}}: A Sample Article Using IEEEtran.cls for IEEE Journals}

% \IEEEpubid{0000--0000/00\$00.00~\copyright~2021 IEEE}
% Remember, if you use this, you must call \IEEEpubidadjcol in the second
% column for its text to clear the IEEEpubid mark.

\maketitle

\begin{abstract}
 Recent aerial object detection models rely on a large amount of labeled training data, which requires unaffordable manual labeling costs in large aerial scenes with dense objects. Active learning effectively reduces the data labeling cost by selectively querying the informative and representative unlabelled samples. However, existing active learning methods are mainly with class-balanced settings and image-based querying for generic object detection tasks, which are less applicable to aerial object detection scenarios due to the long-tailed class distribution and dense small objects in aerial scenes. In this paper, we propose a novel active learning method for cost-effective aerial object detection. Specifically, both object-level and image-level informativeness are considered in the object selection to refrain from redundant and myopic querying. Besides, an easy-to-use class-balancing criterion is incorporated to favor the minority objects to alleviate the long-tailed class distribution problem in model training. We further devise a training loss to mine the latent knowledge in the unlabeled image regions. Extensive experiments are conducted on the DOTA-v1.0 and DOTA-v2.0 benchmarks to validate the effectiveness of the proposed method. For the ReDet, KLD, and SASM detectors on the DOTA-v2.0 dataset, the results show that our proposed MUS-CDB method can save nearly 75\% of the labeling cost while achieving comparable performance to other active learning methods in terms of mAP. \href{https://github.com/ZJW700/MUS-CDB}{\textit{Code}} is publicly online.
\end{abstract}

 %----------------------------------------------------------
\section{Introduction}
\begin{figure}[!t]
\centering
\includegraphics[width=0.49\textwidth]{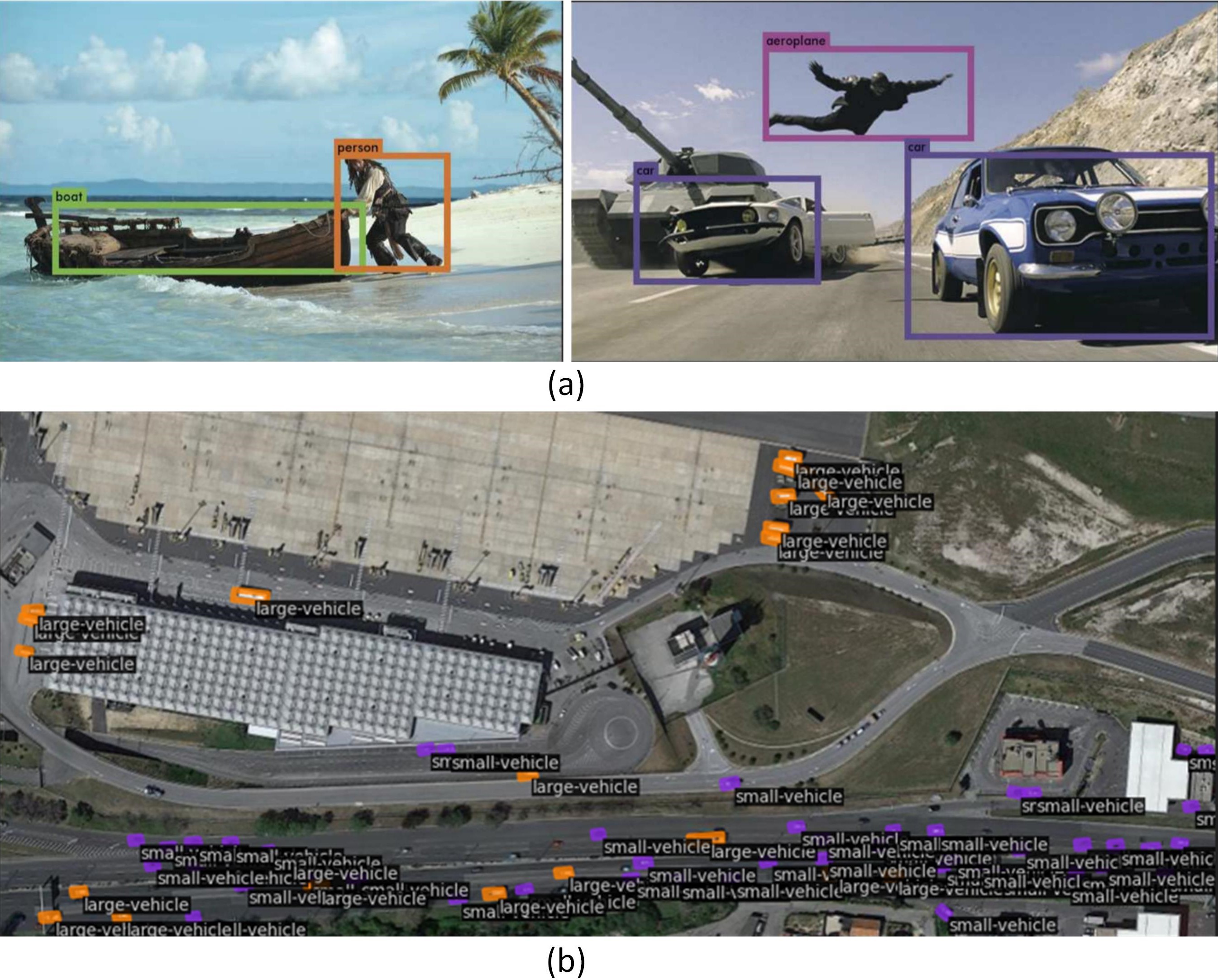}\caption{(a) Generic object detection: objects  are usually sparsely distributed. (b) Aerial object detection: objects in remote sensing images are usually small, blurred, and densely distributed with complex backgrounds.}
\label{intro}
\end{figure}

\IEEEPARstart{A}{erial} object detection has received much attention in recent years due to its important role in land and resources survey, mapping in Geographic Information Systems (GIS), and other fields field~\cite{8113128}. However, existing aerial object detectors usually require a large amount of training data with an expensive bounding-box annotations process~\cite{hua2020relation,zhang2022artificial, }. Active learning~(AL) is a machine learning technique that selectively queries the informative unlabeled examples for annotation to reduce the annotation cost. It has been successfully applied to the generic object detection for efficient annotation~\cite{kao2018localization, tang2021qbox, desai2020towards, roy2018deep}. However, existing active learning methods can hardly be applied to remote-sensing images. As shown in Figure~\ref{intro}, the objects in aerial remote sensing images are usually small, blurred, and densely distributed in the complex background~\cite{wang2022hybrid}. The existing AL methods do not sufficiently consider such characteristics. 
%We strive to design an effective active learning method for aerial object detection that  takes into account the challenges of remote sensing images.

% Active learning for aerial object detection mainly studies effective query strategies and types. 
Two aspects need to be considered in active object detection, i.e., \textbf{query strategy} and \textbf{query type}.
The former investigates the measurements of the informativeness of the data, and the latter designs an efficient manner {color{red} of acquiring} knowledge from the oracle. 

For the query strategy, most existing AL methods evaluate the uncertainty criteria of the unlabeled data. However, these standard criteria neglect one of the most notable problems of aerial remote sensing data -- class imbalance \cite{li2017cost}. As a result, query by uncertainty may intensify the imbalance problem and challenge the model training. The uncertain samples tend to come from  the classes with rare samples, and class preference should be considered in aerial object detection. However, introducing such an AL criterion is challenging, according to the literature~\cite{ning2022active}, due to the difficulty of predicting labels with limited training data.

\begin{figure}[!t]
\centering
\includegraphics[width=0.49\textwidth]{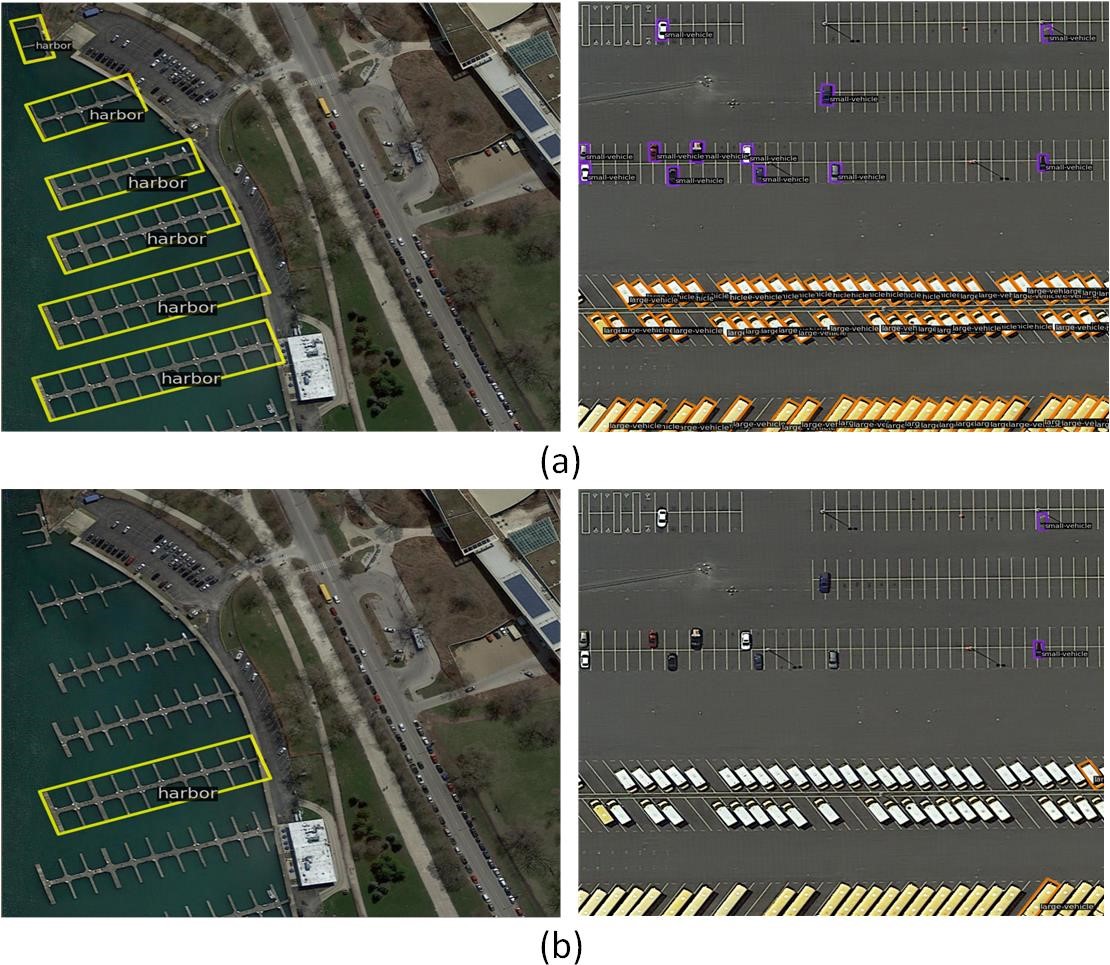}
\caption{Illustration of different types of AL methods. (a) Image-based sampling annotates the whole image with inefficient and redundant annotation. (b) Object-based sampling selects objects that the model is most uncertain.}
\label{query type}
\end{figure}

For the query type, existing solutions can be roughly divided into two categories, image-based (shown in Figure~\ref{query type} (a)), and object-based (shown in Figure~\ref{query type} (b)). Image-based AL methods estimate the uncertainty of the whole scene and require the bounding-box annotation of all objects in the scene \cite{xu2021adaptive,xu2020using}. 
%Although they are adept at capturing the comprehensive contextual information of the image, 
These approaches suffer from inefficient and redundant labeling problems as an aerial image usually contains many similar objects.
%with information redundancy, annotating all of them may lead to a waste of labeling costs. 
Object-based AL methods~\cite{tang2021qbox,desai2020towards} query a specific object (i.e., bounding box) rather than the whole image for more fine-grained and cost-effective annotation.
%, which are more suitable to the aerial data. 
However, 
%these approaches also confront limitations. On the one hand, 
%they ignore the contextual information of the objects, i.e., 
They only consider the uncertainty of the object but neglect the spatial information and the semantic structure of the image. On the other hand, they may introduce training noise as each image is partially annotated with particular objects,  
%only the supervision of particular objects can be obtained, 
and the remaining regions are treated as background with remaining unlabeled objects, which would mislead the model training.
%due to the latent foreground objects.

To address the problems above, in this paper, we present a novel active learning method for aerial object detection -- Mixed Uncertainty Sampling with Class Distribution Balancing~(MUS-CDB). 
For the query type, we propose an object-based mixed uncertainty sampling (MUS) method to label the most informative and representative objects. 
Unlike existing work, MUS addresses the limitations in object-based and image-based methods, i.e., the redundant and myopic information querying, by considering both object-level and image-level uncertain cues. 
For the query strategy, we propose a novel sample selection criteria with class distribution balancing (CDB), identifying the most helpful object samples to improve the performance of the current detection model while balancing the class distribution of training samples further to enhance the model's capability on rare classes. 
We propose an effective training scheme associated with a loss function, which  effectively mines the latent knowledge in the unlabeled regions that have not been queried as positive samples.  
Extensive experiments on DOTA-v1.0~\cite{xia2018dota} and DOTA-v2.0~\cite{ding2021object} show that the proposed method can significantly outperform the conventional image- and object-based active learning methods for aerial object detection.

We summarize our contributions as the following:
\begin{enumerate}
\item{We propose a novel active learning method for aerial object detection, which considers the characteristics of remote sensing data to actively select informative and representative object samples.}
\item{We propose a training scheme associated with a loss function for querying with partially labeled data. It robustly exploits the queried information from partial labels and effectively improves the model's capability.}
\item{Extensive experiments on DOTA-v1.0 and DOTA-v2.0 validate the effectiveness and practicability of the proposed method for aerial object detection.}
\end{enumerate}

The rest of this paper is structured as follows. In Section~\ref{related-work}, we discuss the related work. In Section~\ref{Methodology}, we introduce the proposed method in detail. The experimental results are presented in Section~\ref{Experiments}. Section~\ref{Conclusion} is the conclusion.

%-----------------------------------------------
\section{RELATED WORK}  \label{related-work}
\subsection{Aerial Object Detection}
% 旋转目标检测
%Aerial object detection is an essential task in computer vision \cite{8113128}. 
Compared with generic object detection, aerial object detection is more challenging.
%for the following reasons. Firstly, the objects in remote sensing images have various orientations and aspect ratios. The detector often needs to predict the angle of the bounding box in order to locate the object accurately. Secondly, the objects in remote sensing images are usually small, and their visual features are easily affected by complex imaging processes, noise, and occlusion. 
Many works have been proposed to solve the specific problems of aerial object detection.

Some methods use multiple anchors with different angles, scales, and aspect ratios for regressing the bounding boxes \cite{ma2018arbitrary, zhang2018toward}. However, these methods are usually computationally expensive.
Another group of methods uses only horizontal anchors to detect rotated objects to improve efficiency. Ding et al. \cite{ding2019learning} propose an RoI Transformer to convert Horizontal RoIs (HRoIs) into RRoIs, which reduces the number of anchors. DRN~\cite{pan2020dynamic} performs orientated object detection through dynamic feature selection and optimization. Wei et al. \cite{wei2022learning} propose a vision-transformer-based feature-calibrated guidance (CG) scheme to enhance feature channel communications. CSL \cite{yang2020arbitrary} treats angle prediction as a classification task to avoid discontinuous boundary problems. ReDet \cite{han2021redet} is committed to improving the model's feature representation by modifying the backbone to generate rotation-equivariant features and employing RiRoI Align in the detection head to extract rotation-invariant features. {Based on the popular Rotate-YOLO method, Sun et al. \cite{sun2021bifa} introduced a bi-directional feature fusion module and an angle classification technique to a YOLO-based rotated ship detector.}

% 小目标检测方法
For the challenge of small objects, many works address the challenge by designing networks that handle scale diversity. For instance, Qiu et al. \cite{qiu2019a2rmnet} propose an adaptive aspect ratio network. It assigns different weights to the feature maps and predicts proper ratio aspects for different objects. Li et al.~\cite{li2019deep} design a perceptual generative adversarial network to enhance feature response, which converts the features of small targets into large ones with homogeneous attributes. Zheng et al.~\cite{zheng2020hynet} propose a hyper-scale detector to learn scale-invariant representations of objects. Liang et al. \cite{liang2021anchor} present a dynamic enhancement anchor network (DEA-Net) that contains a sample generator to augment the training data and a discriminator to distinguish the samples between an anchor-based unit and an anchor-free unit to help train the detector. Liu et al.~\cite{liu2021abnet} design an enhanced effective channel attention (EECA) mechanism, an adaptive feature pyramid network (AFPN), and a context enhancement module (CEM) in an adaptive balanced network (ABNet) to capture more discriminative features.
%There are also methods that utilize anchor-free approaches to detect rotated objects.
{Liu et al. \cite{liu2022single} present a method to balance accuracy and speed by using a one-stage detector for high efficiency and design a balanced feature pyramid network (BFPN) that adaptively balances semantic and spatial information between high-level and low-level features.} Most existing methods are based on the assumption that a large number of labeled data for model training is available. In fact, it is usually unavailable in real remote sensing applications due to the prohibitive cost of annotating large-scale remote sensing images with massive objects. 

\subsection{Active Learning in Object Detection}
% 介绍AL（定性，当前任务定性，分类，延伸）
% Active learning aims to reduce the labeling cost~\cite{settles2009active}. 
 {
 The primary objective of active learning is to train a model with maximum accuracy while minimizing the number of labeled samples required for training~\cite{settles2009active}. This is achieved by selecting the most informative samples for labeling using various selection methods.}
 %, such as those based on uncertainty and representativeness. 
 %Despite the reduced number of labeled samples, AL still necessitates expert labeling of the selected samples. 
 %Therefore, active learning cannot completely eliminate labeling costs but instead reduces them by selecting the most valuable samples for labeling. 
 %This approach efficiently utilizes expert labeling resources while achieving high model accuracy.}
 It has been applied to many important tasks in computer vision, such as classification~\cite{sinha2019variational, ebrahimi2020minimax, beluch2018power}, detection~\cite{roy2018deep, aghdam2019active, yoo2019learning} and segmentation~\cite{cai2021revisiting, casanova2020reinforced, kasarla2019region}. 
 %There are two settings in active learning literature, one is the membership query, and the other is the pool-based setting. 
 %Among them, the membership query method directly generates informative samples from the feature space for querying  \cite{mahapatra2018efficient, mayer2020adversarial, zhu2017generative}. In contrast, the pool-based active learning methods assume a large pool of unlabeled data for sampling. They try to select the most informative examples from the pool for querying. In this paper, we mainly study the latter setting. 

% For the query strategies(定性，分类，介绍)
Active learning mainly studies effective query strategies and query types. Many selection criteria are proposed for the former research topic to evaluate the informativeness of the unlabeled data from different aspects. Existing methods can mainly be divided into the uncertainty-based approaches~\cite{yan2018cost, kirsch2019batchbald, yoo2019learning, roy2018deep, kao2018localization}, representativeness-based approaches~\cite{sener2017active, sinha2019variational} and hybrid approaches~\cite{huang2010active, tang2019self, ash2019deep, agarwal2020contextual}. Uncertainty-based methods prefer the examples with high prediction uncertainty of the model. For example, Yoo et al.~\cite{yoo2019learning} add a loss prediction branch to the neural network to predict the loss of unlabeled samples. The data with large predicted loss will be queried from the oracle. Roy et al. \cite{roy2018deep} estimate the uncertainty by the difference between the convolutional layers of the object detector backbone. The data with high divergence are preferred in active selection.  Kao et al. \cite{kao2018localization} propose ``localization tightness” and ``localization stability” criteria. They measure the overlap ratio between region proposals and final predictions and the prediction changes of the original and corrupted images to evaluate the uncertainty.  
% Weiping et al. \cite{yu2022consistency}  leverage mutual information to encourage a balanced class distribution in active selection.
 Representativeness-based methods prefer data that can represent the latent data distribution. Sener et al. \cite{sener2017active} propose a Coreset method to query the data that can cover the whole dataset with a minimum radius. Sinha et al. \cite{sinha2019variational} train a variational autoencoder (VAE) and an adversarial network to classify the unlabeled and labeled data, then select the predicted data from the unlabeled one. For the hybrid methods, Ash et al. \cite{ash2019deep} cluster the gradient of the target model's final output layer as the feature of the unlabeled samples that contain the uncertainty information. Sharat et al. \cite{agarwal2020contextual} use the probability vector predicted by CNN to select objects located in different backgrounds.

For the query type, many methods are proposed to improve the querying efficiency by an effective query type, i.e., taking an object rather than the image as the basic unit for querying~\cite{wang2018towards, tang2021qbox, desai2020towards}. For example, Tang et al.~\cite{tang2021qbox} consider the partial transfer object detection task and query the source objects which are informative and transferrable to the target domain. Laielli et al. \cite{laielli2021region} propose a region-level sampling method, which calculates the score of each region according to an accumulation of the informativeness and similarity of each query-neighbor pairing within a region and finally selects the image region with the highest score to send to the experts for annotation. Xie et al. \cite{xie2022towards} propose a region-based active learning approach for semantic segmentation with domain shift. The authors evaluate the informativeness by the category diversity of pixels within a region and the classification uncertainty. Liang et al. \cite{liang2022exploring} propose a sampling method combining spatial and temporal diversity to label the most informative frames and objects in multimodal data.
%Most existing AL work are less applicable to aerial object detection due to the neglect of the challenges of this task. 

\begin{figure*}[!t]
\centering
\includegraphics[width=.9\textwidth]{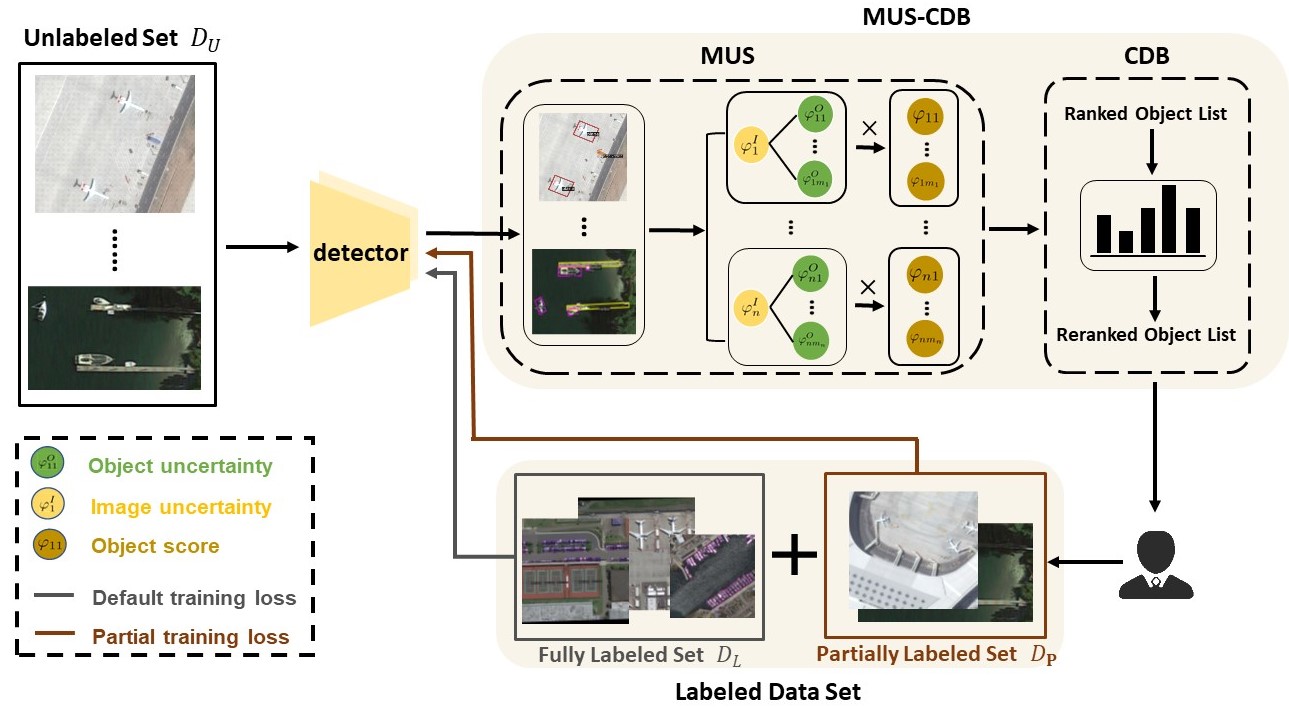}\caption{Overall framework of the proposed method. The overall framework of the proposed method consists of two modules: Mixed Uncertainty Sampling (MUS) and Class Distribution Balancing (CDB), which together form the active learning sampling strategy. The MUS module combines image uncertainty and object uncertainty to obtain the final object information. The CDB module performs class-balanced sampling on the objects selected by MUS based on the prior distribution of labeled objects in the pool. We take the model prediction object as the basic sampling unit and send it to experts for labeling.}
\label{framework}
\end{figure*}

\section{The Methodology}\label{Methodology}

\subsection{Problem Definition}
{The problem definition for remote sensing object detection using active learning is to reduce the labeling cost by selecting informative samples from a large unlabeled dataset $\mathcal{D}_U$ to train a well-performing detector $f$. The problem is defined by three sets of data: a small fully labeled set $\mathcal{D}_L$ used to initialize the model, a large unlabeled set $\mathcal{D}_U$ used for data selection, and a partially labeled set $\mathcal{D}_P$ updated with informative queries selected by an active learning method. $\mathcal{D}_L$ is expressed as 
\begin{equation}
\mathcal{D}_L = \{(X_i^L, \mathcal{Y}_i^L)\}_{i=1}^{N_L},
\end{equation}
where $X_i^L$ represents the $i^{\operatorname{th}}$ image in the fully labeled set $\mathcal{D}_L$, and
$\mathcal{Y}_i^L=\{(\textbf{c}_{ij}^L, \textbf{b}_{ij}^L)\}_{j=1}^{n_i}$
with $\textbf{c}_{ij}^L$ being the one-hot class label belonging to one of the $C$ known classes in label space and $\textbf{b}_{ij}^L$ being the bounding box label. In summary, $\mathcal{D}_L$ is a set of $N_L$ images, each annotated with one or more ground truth bounding boxes, where each bounding box is associated with a one-hot class label. $\mathcal{D}_U$ is expressed as 
\begin{equation}
\mathcal{D}_U = \{(X_i^U)\}_{i=1}^{N_U}
\end{equation}
where $N_U \gg N_L$, and $\mathcal{P}_U$ is the model prediction on the unlabeled set $\mathcal{D}_U$, expressed as
\begin{equation}
\mathcal{P}_U = \{(X_i^U, \mathcal{\hat{Y}}_i^U)\}_{i=1}^{N_U}
\end{equation}
where
$\mathcal{\hat{Y}}_i^U = \{(\mathbf{\hat{c}}_{ij}^U, \mathbf{\hat{b}}_{ij}^U)\}_{j=1}^{\hat n_i}$
 denotes the model's prediction. Specifically, $\mathcal{\hat{Y}}_i^U$ consists of two parts, $\mathbf{\hat{c}}_{ij}^U$ is the predicted class probability vector from the classification head where
$\mathbf{\hat{c}}_{ij}^U = \{\hat{c}_{ijk}^U\}_{k=1}^C$,
and $\mathbf{\hat{b}}_{ij}^U$ is the predicted bounding box position from the regression head.

To train a well-performing detector $f$ with minimal labeling effort from $\mathcal{D}_U$, a sampling function ($\varphi$) is employed to select informative samples from the unlabeled image set $\mathcal{D}_U$. Specifically, the sampling function $\varphi$ takes $\mathcal{P}_U$ as input to select the informative data for labeling. For the proposed object-based active learning method, we evaluate each predicted bounding box in $\mathcal{P}_U$ and select the top $N$ most informative bounding boxes for labeling. Once labeled, these bounding boxes are added to the partially labeled set 
\begin{equation}
\mathcal{D}_P = \{(X_i^P, \mathcal{Y}_i^P)\}_{i=1}^{N_P} 
\end{equation}
to improve the object detector $f$.}

In this work, we design two modules to perform a cost-effective selection for objects, i.e., the object-based mixed uncertainty sampling module (MUS) and the class distribution balancing module (CDB). {The overall framework of the method is shown in Figure \ref{framework}.} Next, we will introduce the proposed method in detail and elaborate on the training scheme to fully utilize the partial labels.

%\subsection{Selection Strategy}
\subsection{Mixed Uncertainty Sampling}
As aforementioned, existing object-based sampling methods mainly consider the information of the prediction box itself, i.e., category uncertainty or regression uncertainty, but neglect the spatial information and the semantic structure of the image. To tackle this problem, we propose to consider both uncertainty of the image and the object for more comprehensive data evaluation, which incorporates both global and local information. 

As for the image uncertainty, if there are many predicted objects with high uncertainty in an image, this image should be preferred to be selected as a labeling candidate. To this end, we evaluate and aggregate the uncertainty values with the most confident model predictions (i.e., objects whose predicted class confidence is greater than a specific threshold) to indicate the uncertainty value of the whole image. Specifically, the image uncertainty $\varphi_i^I$ for a given image $X_i^U$ is formulated as
\begin{equation}
\begin{aligned}
   \varphi_i^I = 1 -  \frac{1}{|\mathcal{S}^{\theta}_i|}\sum_{j \in \mathcal{S}^{\theta}_i} \max \mathbf{\hat{c}}_{ij}^U \,,
\end{aligned}
\label{eq:image}
\end{equation}
where 
       $\mathcal{S}^{\theta}_i = \{ j | \max \mathbf{\hat{c}}_{ij}^U>\theta , \forall j = 1,\dots,\hat n_i \}$
       ,
and $|\cdot|$ represents the number of elements in the set and $\theta$ is the threshold. {The image uncertainty value $\varphi_i^I$ is calculated as the difference between 1 and the average confidence score of the predicted bounding boxes. Only predicted bounding boxes with class confidence scores greater than the threshold $\theta$ are used to calculate the average confidence score. The set ${S}^{\theta}_i$ represents the indices of the predicted bounding boxes in an image $i$ whose maximum class confidence score is greater than the threshold $\theta$. The value of $\varphi_i^I$ is high when an image has many predicted bounding boxes with low confidence scores, which may occur when an image contains many objects that are difficult to distinguish, resulting in inconsistent and low-confidence predictions. Therefore images with higher values of $\varphi_i^I$ are more likely to contain informative knowledge of rare patterns, making them more suitable for selection.}
%One phenomenon to imply this motivation is that the distribution of different categories of remote sensing images has certain regularity. Some head categories, such as [small vehicle] and [large vehicle], are densely distributed and very similar to each other; Some medium categories, such as [bridge], are sparsely distributed in the picture, and the background of the picture is blurry and difficult to distinguish; Tail categories, such as [helicopter], have a small number and poor performance. On the one hand, these images with similar and densely distributed objects may introduce information redundancy; on the other hand, they may intensify the class imbalance problem in model learning. For these reasons, we should avoid querying these images. 

To consider the object-level information in the querying, we employ entropy to evaluate the uncertainty of each predicted bounding box. Specifically, the object uncertainty $\varphi_{ij}^O$ is calculated as follows
\begin{equation}
\begin{aligned}
   \varphi_{ij}^O = -\sum_{k=1}^C \mathbb P(\hat c_{ijk}^U|X_i^U) \log \mathbb P(\hat c_{ijk}^U|X_i^U)\,,
\end{aligned}
\label{eq:object_uncertainty}
\end{equation}
where $\mathbb P(\hat c_{ijk}^U|X_i^U)$ is the predicted probability of the $j^{\operatorname{th}}$ bounding box in the image $X_i^U$ on class $k$. 

Next, we combine the image uncertainty $\varphi_i^I$ and object uncertainty $\varphi_{ij}^O$ to obtain the final object information score $\varphi_{ij}$, as shown in Equ.~\eqref{object_score}. 
\begin{equation}
    \begin{aligned}
    &\varphi_{ij} = \varphi_i^I \cdot \varphi_{ij}^O\,, \\
    \forall i=1,&\dots,N^U,\; \forall j=1,\dots,\hat n_i\,.
    \end{aligned}
    \label{object_score}
\end{equation}

%We note that this criterion is combined with the image uncertainty and class balancing to conduct active selection. The overall query strategy reasonably considers the spatial information, the semantic structure of the image, and the instance-level uncertainty. We will show the effectiveness of the selection criteria empirically in the experiments.

\subsection{Class Distribution Balancing}
\begin{figure}[!t]
\centering
\includegraphics[width=0.5\textwidth]{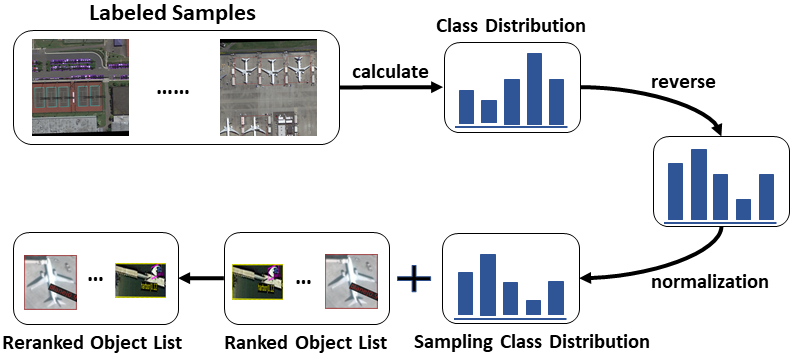}
\caption{The proposed Class Distribution Balancing module (CDB) procedure. We reverse the class distribution of the labeled samples and perform a softmax operation to obtain the distribution that the class should satisfy when sampling.}
\label{CB}
\end{figure}

Remote sensing data suffer from the problem of class imbalance~\cite{xia2018dota}, as some categories are naturally rare. This phenomenon will significantly jeopardize the model performance, especially for the rare categories. To tackle this problem, we propose a criterion emphasizing the categories with low occurrence frequency in active querying. {The process of class distribution balancing is shown in Figure \ref{CB}.} Specifically, we first identify the rare categories by counting the objects of each class on the labeled set, which is a combination of the initial fully labeled set $\mathcal{D}_L$ and the partially labeled set $\mathcal{D}_P$. Let $a_k$ denote the number of objects in class $k$, $\forall k=1,\dots, C$. We aim to query more objects in classes with fewer objects by imposing a preference $\zeta_k$, which is inversely proportional to $a_k$, on each class during the selection phase. Specifically, we calculate the preference value $\zeta_k$ as follows: 
\begin{equation}
\begin{aligned}
    \zeta_k = \frac{exp(\beta_k)}{\sum\limits_k exp(\beta_k)}\,
\end{aligned}
\label{eq:distribution}
\end{equation}  
where
\begin{equation}
\begin{aligned}
    \beta_k = 1-\frac{a_k}{\sum\limits_{k}a_k}\,.
\end{aligned}
\end{equation}
we first calculate the distribution probability of each class in the labeled set based on $a_k$, then take the inverse to obtain the class weight $\beta_k$ during sampling.
We then use the softmax function, as shown in Equ.~\eqref{eq:distribution}, to compute the expected class distribution during sampling. This allows us to set preferences for different classes and selectively query objects in classes with fewer objects during the selection phase, thereby improving the performance and accuracy of the model.

%{\color{red} Sections A and B provide a detailed explanation of MUS-CDB, an object-based two-stage active learning sampling method. 
%We describe the steps involved in Mixed Uncertainty Sampling (MUS) and Class Distribution Balancing (CDB) to select high-information objects for annotation. 
%Section D summarizes the entire sampling process of MUS-CDB, where we select uncertain samples from the unlabeled set $\mathcal{D}_U$.} 

\subsection{{Two-Stage Sampling}}
{We summarize our active learning method in Figure~\ref{framework} and Algorithm\ref{alg}, which aims to select high-information objects for annotation while optimizing the labeling budget and improving the model's performance. 
The method includes mixed uncertainty sampling (MUS) and Class Distribution Balancing (CDB) stages. 
In the MUS stage, as described in Figure~\ref{framework}, we take the unlabeled images from the $\mathcal{D}_U$ and feed them into a pre-trained object detector to obtain the model prediction. 
We then calculate the uncertainty value of each predicted object using Equ.~\eqref{eq:object_uncertainty} and the uncertainty value of each image using Equ.~\eqref{eq:image} based on the model's output class probability vector $\mathbf{\hat{c}}_{ij}^U$. Finally, we calculate the information value of each predicted object using Equ.~\eqref{object_score}. 
In the CDB stage, we allocate the labeling budget for each class based on the corresponding class preference value $\zeta_k$. Specifically, we determine the labeling budget for each class by multiplying the total labeling budget by the class preference value. Then, During the sampling phase, we select high-value objects for annotation based on a balanced class distribution strategy. Specifically, we select enough objects from each class based on their information value in descending order and the labeling budget determined in the previous step. 
If the labeling budget for a specific class has been exhausted, we discard any remaining objects predicted as belonging to that class. Using this approach, we can effectively select high-information objects for annotation while optimizing the labeling budget and improving the model's performance. 
%For more details on the method, please refer to Figure~\ref{framework} and Algorithm\ref{alg}.
}

\subsection{Dealing with Partially Labeled Data}
{To deal with the situations where some samples are fully labeled while others are only partially labeled during the model training, we use different training losses for the two sets. 
For the fully labeled dataset, we use the default training loss of the specific detector; 
For the partially labeled dataset, we use a loss function to effectively mine the latent knowledge in the unlabeled image regions that have not been queried.  
Since some objects in the partially labeled images are not labeled, training the model with these images can mislead the model's classification head. 
To address this issue, we propose an adaptive weight loss function that uses the predicted box background score as the weight for each negative box's classification loss.  
Specifically, in object detection models (\cite{han2021redet, xie2021oriented, yang2021learning, hou2022shape} ), the bounding-box head loss $\mathcal{L}_{bbox}$ is a key component comprising both classification and regression losses. 
The classification loss $\mathcal{L}_{cls}$ measures the accuracy of object classification, including positive and negative sample losses, while the regression loss $\mathcal{L}_{reg}$ measures the accuracy of the predicted bounding box location, consisting only of positive sample losses. 
However, training the model with partially labeled images can introduce noise to the negative sample losses in the classification loss since some objects in the images may not be labeled and are treated as negative samples. 
To address this issue, we propose an adaptive weight loss function for the negative sample losses in the classification loss. Specifically, we add an adaptive weight to each negative sample's classification loss based on the predicted background score of the sample. 
This method can effectively suppress the classification loss of negative samples with low background scores, which the model considers foreground objects.} 
Formally, the $\mathcal{L}_{bbox}$ is defined as follows

% lambda:归一化项
\begin{equation}
\begin{aligned}
    \mathcal{L}_{bbox} = & \mathcal{L}_{cls} + \mathcal{L}_{reg} \\
    &=(\lambda_{cls}\sum_{j=1}^{W}\sum_{k=1}^{C+1}\mathbb I_{i}^{partially}\omega[-\mathbf{{o}_{ijk}}\log (\mathbf{\hat{o}_{ijk}})] + \\
     &\lambda_{cls}\sum_{j=1}^{W}\sum_{k=1}^{C+1}\mathbb I_{i}^{fully}[-\mathbf{{o}_{ijk}}\log (\mathbf{\hat{o}_{ijk}})])  \\
    & + \lambda_{reg}\sum_{j=1}^{W}\sum_{u=1}^{5}\mathbb I_{ij}^{obj}\ell_{s}(\hat{v}_{iju}-v_{iju}) \,,
\end{aligned}
\label{eq:lose}
\end{equation}
where

\begin{equation}
\omega = \left\{
	\begin{aligned}
	1 \quad \mathbb I_{ij}^{obj} = 1\\
	\mu \quad \mathbb I_{ij}^{obj} = 0
	\end{aligned}
	\right
	.
\label{eq:w}
\end{equation}

and 

\begin{equation}
\ell_{s}(x) = \begin{cases}
    0.5(x)^2, \quad &if |x|<1\\
    |x|-0.5, \quad &otherwise
    \end{cases}
    .
\end{equation}

Here $i$ and $j$ are the images and region proposals indexes in a mini-batch, respectively. 
$W$ indicates the number of proposals involved in bounding-box heads in training. 
$\mathbb I_{i}^{partially}$ and $\mathbb I_{i}^{fully}$ are indicator functions. 
When an image $i$ is partially labeled, $\mathbb I_{i}^{partially}$ equals 1 and $\mathbb I_{i}^{fully}$ equals 0. When an image $i$ is fully labeled, $\mathbb I_{i}^{partially}$ equals 0 and $\mathbb I_{i}^{fully}$ equals 1. 
$\mathbb I_{ij}^{obj}$ indicates whether the proposal contains a foreground object. 
$\omega$ is introduced to down-weighting the background objects for robust learning. 
{The classification and regression losses are normalized using the balancing parameters $\lambda_{cls}$ and $\lambda_{reg}$, respectively~\cite{ren2015faster}. 
Specifically, the number of proposals in a mini-batch is used to normalize the classification loss, while the number of positive proposals in the mini-batch is used to normalize the regression loss.}

The $\mathcal{L}_{bbox}$ contains the classification loss (the first two terms) and the box regression loss (the last term). 
For the classification loss, the model predicts a discrete probability distribution over $C$ categories plus the background for each proposal, i.e., $\mathbf{\hat{o}_{ijk}}, \forall k=1, \dots, C+1$. The regression loss is defined over the bounding-box coordinate offsets plus the angle, i.e., $\hat{v}_{iju},\,\forall u=1,\dots,5$. Further, the smooth L1 regularization $\ell_s$ is employed to stabilize the training.

\begin{algorithm}[t]
    \renewcommand{\algorithmicrequire}{\textbf{Input:}}
    \renewcommand{\algorithmicensure}{\textbf{Output:}}
    \caption{Our proposed MUS-CDB}
    \label{alg}
    \begin{algorithmic}[1]
        \Require initial labeled set $D_L$, unlabeled set $D_U$, partailly labeled set $D_P$, annotation budget $N$, detection model $f$.
        \State Train the detection model $f$ with $D_L$.
        \State $\mathcal P^U \leftarrow \{(X_i^U, f(X_i^U) \,|\, \forall i=1,...,N_U\}$.
        \State Remove the objects in $\mathcal P^U$, which highly overlap with the labeled object.
        \State $\mathbf \zeta \gets \{\zeta_k\,|\, \forall k=1,\dots,C\}$.
        \Comment{by Equ.~\eqref{eq:distribution}}
        \State $\Phi \gets \{\varphi_{ij}\,|\,\forall i=1,\dots,N^U,\; \forall j=1,\dots,\hat n_i\}$.
        \Comment{By Equ.~\eqref{object_score}}
        \State $\mathbf {a} \gets \{a_k \,| \,a_k=N\cdot \zeta_k,\, \forall k=1,\dots,C\}$.  \Comment{calculate the labeling budget for each class}
        \State $\Phi \gets \operatorname{sort}(\Phi)$. \Comment{sort $\Phi$ in descending order}
        \While{$N > 0$}
            \State $i, j \gets \{i,j \,|\,\varphi_{ij}=\operatorname{pop}(\Phi)\} $.
            \State $c \gets \text{argmax}\{ \hat{\mathbf c}_{ij}^U\}$.
            \If{$a_c>0$}
                \State $D_P \gets D_P \cup \operatorname{label}(\varphi_{ij})$.  \Comment{label the $\varphi_{ij}$ instance according to the labeling rule}
                \State $a_c \gets a_c - 1$.
                \State $N \gets N - 1$.
            \EndIf
        \EndWhile
        \Ensure the updated queried set $D_P$
    \end{algorithmic}
\end{algorithm}

\section{Experiments}\label{Experiments}

\subsection{Experimental Settings}
\subsubsection{Dataset}
DOTA is a popular object detection dataset in aerial imagery. 
We conduct experiments on DOTA-v1.0~\cite{xia2018dota} and DOTA-v2.0 \cite{ding2021object}. 
Specifically, DOTA-v1.0 contains 2,806 large-scale aerial images and 188,282 objects. DOTA-v2.0 collects more Google Earth, GF-2 Satellite, and aerial images. There are 11,268 images and 1,793,658 objects in DOTA-v2.0. 
Each image is between 800 $\times$ 800 and 20000 $\times$ 20000 pixels in size. 
There are a total of 15 classes in the DOTA-v1.0 [containing plane] (PL), [baseball diamond] (BD), [bridge] (BR), [ground track field] (GTF), [small vehicle] (SV), [large vehicle] (LV), [ship] (SH), [tennis court] (TC), [basketball court] (BC), [storage tank] (ST), [soccer-ball field] (SBF), [roundabout] (RA), [harbor] (HA), [swimming pool] (SP) and [helicopter] (HC). 
DOTA-v2.0 further adds the new categories: [container crane] (CC), [airport] (AP), and [helipad] (HP).

For DOTA-v1.0,  the training and testing sets follow the setup of the previous work \cite{han2021redet}. 
For DOTA-v2.0, since the server for uploading the results is unstable, we use the training set for training and the validation set for testing. 
We crop the original image to 1024 $\times$ 1024 patches with an overlap of 200. 
The random horizontal flip is employed in data augmentation.

\subsubsection{Comparison AL Methods}
We compare the proposed method with various types of active learning methods, including random selection (Random), uncertainty-based active learning methods (Entropy~\cite{settles2009active}), representation-based active learning methods (Coreset~\cite{sener2017active}), detection-specific active learning methods (Localization Stability \cite{kao2018localization}) and object-level active learning methods {(Qbox~\cite{tang2021qbox}, BiB \cite{vo2022active} and TALISMAN \cite{kothawade2022talisman}.} 
%We introduce two baseline methods, namely image-based and object-based query type{\color{red}, which aim to acquire informative samples by selecting informative images and objects, respectively}. 
{Both Qbox~\cite{tang2021qbox} and our proposed method utilize an object-based query approach to select informative samples for annotation. In contrast, other methods employ an image-based query approach. 
In Qbox~\cite{tang2021qbox}, we use the model-predicted bounding boxes as candidate objects for sampling and calculate their informativeness using the inconsistency metric defined in Qbox~\cite{tang2021qbox}. 
The predicted bounding boxes are then sent to the oracle for labeling in descending order of informativeness. 
%We also use an object-based query approach for our proposed method but employ a different informativeness metric. 
Meanwhile, for image-based methods such as entropy~\cite{settles2009active}, we calculate the entropy of each object using the model-predicted class probabilities and then average them to obtain an information score for the image. 
For TALISMAN, we set the sampling quantity for rare classes in DOTA-v1.0 and DOTA-v2.0 to 2 in the query set. 
%The rare classes in both datasets are listed in Table \ref{table:cateAP2} and Table \ref{table:cateAP1}.
}

\subsubsection{Labeling Rules} Image-based and object-based AL methods have different labeling rules. 
Object-based AL methods send the predicted bounding box as a unit to the oracle for labeling. 
If there is only one unlabeled ground-truth object in the query bounding box, the label of that object is returned. 
If multiple unlabeled ground-truth objects exist in the query bounding box, calculate the IoU of each ground-truth object and the query bounding box, and return the object's label with the largest IoU. 
After the queried bounding box obtains the object annotations, the annotation cost (i.e., the number of annotated objects) is increased by one. 

\subsubsection{The detectors} The active learning experiments were conducted using the ReDet~\cite{han2021redet} architecture with ReResNet50 pretrained on ImageNet \cite{krizhevsky2017imagenet} as the baseline detector on the DOTA-v1.0 and DOTA-v2.0 datasets. 
{To further validate the generality and effectiveness of our proposed method MUS-CDB, we conducted experiments on other two single-stage remote sensing object detectors KLD \cite{yang2021learning}, SASM \cite{hou2022shape} and one other two-stage remote sensing object detectors Oriented R-CNN~\cite{xie2021oriented}.}

\subsubsection{Dataset Partitioning}
Regarding the dataset partitioning, for DOTA-v1.0, We randomly select 5\% images (1052 images) from the training and validation sets as the initially fully labeled set, with the remaining images making up the unlabeled set. 
In each round of active learning, 5000 objects are selected from the unlabeled set for querying. 
These objects are then labeled (we use the ground truth in the experiments) and added to the partially labeled set for fine-tuning the detector. 
This process is repeated for multiple rounds until the budget for annotation is exhausted.
For DOTA-v2.0, we randomly selected 500 images from the training set as the initially fully labeled set, with the remaining images being considered the unlabeled set. 
In each round, 1000 objects are selected from the unlabeled set for querying.
The model was fine-tuned on the labeled set (combination of fully labeled set and partially labeled set) for a fixed number of epochs in each round of active learning. 
\subsubsection{Training Settings}
For ReDet, the model with default pre-trained weights will be fine-tuned on the labeled set for 12 epochs with a batch size of 2, and the initial learning rate is set to 2.5 $\times$ $10^{-3}$, using the SGD optimizer mentum of 0.9 and weight decay of 1 $\times$ $10^{-4}$. 
The learning rate is reduced by 0.1 at 8 and 11 epochs. 
%{\color{red} The model trained with these settings converged quickly and achieved an expected decrease in the loss value. 
{On the DOTA-v1.0 dataset, the loss value of the model decreased from an initial value of 1.076 to a final value of 0.656, and the entire loss curve showed a downward trend. 
Similarly, on the DOTA v2.0 dataset, the loss value of the model decreased from an initial value of 1.694 to a final value of 0.8551, and the entire loss curve showed a  downward trend. 
For the other two-stage detector Oriented R-CNN~\cite{xie2021oriented}, we set the initial learning rate to 5 $\times$ $10^{-4}$ and keep other variables consistent with ReDet's settings. 
For KLD~\cite{yang2021learning}, we trained the model for 14 epochs with a batch size of 2. 
The initial learning rate was set to 1 $\times$ $10^{-3}$, and the settings for the SGD optimizer were fixed. 
The learning rate was reduced by 0.1 at 10 and 13 epochs. 
Similarly, for SASM~\cite{hou2022shape}, we used the same training strategy as KLD~\cite{yang2021learning}, including the same number of epochs, batch size, learning rate, optimizer, and weight decay.}
\subsubsection{Other Settings}
For active learning, the parameter settings involved in our sampling method are as follows. 
{We implement the adaptive weight $\mu$ in Equ.~\eqref{eq:w} as the background score predicted by the model for proposals.} 
In this way, we can reduce the impact of the partially labeled data by down-weighting the loss of the false-negative objects, which typically have lower background scores than background regions. 
Regarding the hyperparameter $\theta$ in Equ.~\ref{eq:image}, we set it to 0.10, and Table \ref{table:threshold} lists the performance change of our method under different values of this parameter. All experiments are conducted on 4 NVIDIA GeForce GTX 2080Ti GPUs.

\begin{table*}[!htbp]
\caption{ Comparison of mAP(\%) for ReDet~\cite{han2021redet}, Oriented R-CNN~\cite{xie2021oriented}, KLD~\cite{yang2021learning} and SASM~\cite{hou2022shape} using different AL methods on DOTA-v1.0~\cite{xia2018dota} and DOTA-v2.0~\cite{ding2021object} with different numbers of Cycles. 
Numbers in bold are the best results per column. 
* denotes the AL method that follows the object-based query type. 
The Cycle-0 column reports the detector's performance without active learning sampling.}
\label{tab:result}
\raggedright

\begin{subtable}[t]{\textwidth}
\label{table:single-stage}
\centering
\setlength{\tabcolsep}{1.7pt}{
\begin{tabular}{c|ccccc|ccccc|ccccc} 
\hline\hline\noalign{\smallskip}
Detector  & \multicolumn{10}{c|}{ReDet~\cite{han2021redet}}   & \multicolumn{5}{c}{Oriented R-CNN\cite{xie2021oriented}}       \\ 
\noalign{\smallskip}
\hline\noalign{\smallskip}
Dataset  & \multicolumn{5}{c|}{DOTA-v1.0~\cite{xia2018dota}}  & \multicolumn{5}{c|}{DOTA-V2.0~\cite{ding2021object}}  & \multicolumn{5}{c}{DOTA-v2.0~\cite{xia2018dota}} \\ \noalign{\smallskip}
\hline\noalign{\smallskip}
AL cycle                & Cycle-0 & Cycle-1 & Cycle-2 & Cycle-3 & Cycle-4 & Cycle-0 & Cycle-1 & Cycle-2 & Cycle-3 & Cycle-4 & Cycle-0 & Cycle-1 & Cycle-2 & Cycle-3 & Cycle-4 \\ 
\noalign{\smallskip}
\hline\noalign{\smallskip}
Random                  & 53.9    & 58.1    & 60.3    & 61.8    & 64.4    & 22.8    & 24.6    & 25.1    & 25.3    & 26.2    & 6.3     & 6.3     & 9.2     & 10.1    & 10.4  \\
Coreset~\cite{sener2017active}    & 53.9    & 55.8    & 59.6    & 62.1    & 63.6    & 22.8    & 23.5    & 23.8    & 25.1    & 26.6    & 6.3     & 7.5     & 10.2    & 11.5    & 13.0  \\
Localization Stability~\cite{kao2018localization} & 53.9    & 61.6    & 63.7    & 66.8    & 66.9    & 22.8    & 27.1    & 30.9    & 31.1    & 33.5    & 6.3     & 13.9    & 16.9    & 16.0    & 17.1  \\
Entropy~\cite{kao2018localization} & 53.9    & 64.2    & 65.7    & 67.2    & 67.4    & 22.8    & 28.3    & 30.9    & 31.7    & 35.0    & 6.3     & 14.8    & 20.1    & 22.0    & 24.8   \\
Qbox$^*$~\cite{tang2021qbox}  & 53.9    & 58.7    & 61.4    & 64.1    & 65.2    & 22.8    & 24.1    & 27.8    & 28.8    & 31.2     & 6.3     & 14.7    & 17.1    & 18.7    & 20.4    \\
BiB~\cite{vo2022active}   & 53.9    & 60.9    & 65.0    & 66.4    & 66.8    & 22.8    & 23.7    & 24.4    & 26.6    & 28.2    & 6.3     & 11.6    & 16.3    & 18.6    & 19.9   \\
TALISMAN-FLMI~\cite{kothawade2022talisman}    & 53.9    & 58.9    & 60.7    & 61.0    & 62.0    & 22.8    & 25.6    & 31.4    & 33.0    & 36.6    & 6.3     & 13.0    & 17.3    & 19.1    & 20.0  \\
TALISMAN-GCMI~\cite{kothawade2022talisman} & 53.9    & 59.9    & 61.3    & 63.5    & 65.4    & 22.8    & 30.4    & 32.9    & 38.4    & 40.0    & 6.3     & 13.8    & 17.3    & 19.8    & 20.9     \\ 
\noalign{\smallskip}\hline\noalign{\smallskip}
MUS$^{*}$               & 53.9    & 64.6    & 66.3    & 68.3    & 69.1    & 22.8    & 33.0    & 35.8    & 38.0    & 39.5    & 6.3     & 15.9    & 20.7    & 24.7    & 28.2   \\
CDB$^{*}$               & 53.9    & \bf{65.2}    & 66.9    & \bf{69.2}    & 69.2    & 22.8    & 33.2    & 36.1    & 39.1    & 40.6    & 6.3     & 16.7    & \bf{22.9}    & 25.9    & \bf{28.4}  \\
MUS-CDB$^{*}$           & 53.9    & 64.9    & \bf{67.3}   & \bf{69.2}    & \bf{70.1}    & 22.8    & \bf{34.1}    & \bf{37.2}    & \bf{40.2}    & \bf{40.9}    & 6.3     & \bf{17.3}    & 22.7    & \bf{26.0}    & 28.2 \\
\noalign{\smallskip}\hline
\end{tabular}}

\end{subtable}
~% add desired spacing

\vspace{-7pt}

~% add desired spacing
\begin{subtable}[t]{\textwidth}
\centering
\label{tab:sample_4}
\setlength{\tabcolsep}{8.3pt}{
\begin{tabular}{c|ccccc|ccccc} 
\hline\noalign{\smallskip}
Detector       & \multicolumn{5}{c|}{KLD~\cite{yang2021learning}}  & \multicolumn{5}{c}{SASM~\cite{hou2022shape}}            \\ 
\noalign{\smallskip}\hline\noalign{\smallskip}
Dataset       & \multicolumn{10}{c}{DOTA-v2.0~\cite{xia2018dota}}             \\ 
\noalign{\smallskip}\hline\noalign{\smallskip}
AL cycle                & Cycle-0 & Cycle-1 & Cycle-2 & Cycle-3 & Cycle-4 & Cycle-0 & Cycle-1 & Cycle-2 & Cycle-3 & Cycle-4 \\ 
\noalign{\smallskip}\hline\noalign{\smallskip}
Random                  & 1.9     & 2.1     & 3.3     & 2.9     & 3.7     & 11.7    & 13.3    & 11.8    & 14.1    & 16.2        \\
Coreset~\cite{sener2017active}    & 1.9     & 1.1     & 2.2     & 2.4     & 2.2     & 11.7    & 12.2    & 11.2    & 13.6    & 11.1        \\
Localization Stability~\cite{kao2018localization} & 1.9     & 2.2     & 2.7     & 3.9     & 3.5     & 11.7    & 13.6    & 18.2    & 19.3    & 20.9        \\
Entropy~\cite{settles2009active}  & 1.9     & 2.1     & 4.7     & 5.5     & 8.6     & 11.7    & 13.8    & 14.4    & 14.6    & 15.3       \\
\noalign{\smallskip}\hline\noalign{\smallskip}
% MUS$^{*}$        & -       & -       &         &         &         &         &         &         &         &           \\
% CDB$^{*}$        & -       & -       &         &         &         &         &         &         &         &          \\
MUS-CDB$^{*}$           & 1.9     & \bf{9.8}     & \bf{13.4} & \bf{16.8}    & \bf{21.7}    & 11.7    & \bf{18.6}    & \bf{24.8}    & \bf{28.0}      & \bf{30.0}          \\
\noalign{\smallskip}\hline
\hline
\end{tabular}}
\end{subtable}
\end{table*}

\subsection{Performance Comparison}
\subsubsection{AL Performance Comparisons with Different Detectors} We report the performances of all AL methods on the DOTA-v1.0 and DOTA-v2.0 benchmarks to demonstrate that our method can better mine the most informative objects from the unlabeled data pool in remote sensing scenarios. 
The experimental results of the two datasets are shown in Table~\ref{tab:result}. 

For ReDet, we conducted experiments on both DOTA-v1.0 and DOTA-v2.0 datasets. 
For DOTA-v1.0 in Table \ref{tab:result},  MUS-CDB significantly outperforms the other methods at each active learning cycle. 
Specifically, in terms of mAP, when 5000, 10000, 15000, and 20000 objects were sampled, the MUS-CDB surpassed the random method by 6.8, 7, 7.4, and 5.7, and the sub-optimal method by 0.7, 1.6, 2 and 2.7, respectively. 
%This performance boosts prove that our method can accurately identify informative objects. 
For DOTA-v2.0, MUS-CDB shows more impressive improvements over image-based and object-based sampling methods. 
The results in Table \ref{tab:result} demonstrate that in terms of mAP, when 1000, 2000, 3000, and 4000 objects are sampled, the MUS-CDB surpassed the random method by 9.5, 12.1, 14.9, and 14.7, and the sub-optimal method by { 3.7, 4.3, 1.8 and 0.9}, respectively. 
The performance improvement on DOTA-v2.0 is more significant than that on DOTA-v1.0, which shows our approach can better guide low-performing models to sample highly informative objects based on only a small number of training samples.
Furthermore, Table~\ref{tab:result} demonstrates that our method achieves comparable or even better performance than other methods that sample 4000 unlabeled objects (Cycle-4) in DOTA-v2.0 by only querying 1000 unlabeled objects (Cycle-0), except for TALISMAN-gcmi. 
For the ReDet, KLD and SASM detector on the DOTA-v2.0 dataset, the results show that our proposed MUS-CDB method can save nearly 75\% of the labeling cost while achieving comparable performance to other active learning methods in terms of mAP. 

{Based on the results presented in Table \ref{tab:result}, it is evident that Entropy and Localization Stability perform well in both DOTA-v1.0 and DOTA-v2.0 datasets. 
Conversely, Coreset and Random perform poorly, as expected.
This finding highlights that uncertainty-based methods are more effective than representativeness-based ones in dealing with chaotic scenes. 
Moreover, the object-based sampling method Qbox performs poorly on either dataset. 
Qbox was designed for generic scenes, which differ from remote-sensing scenes. 
The sampling strategy of Qbox cannot accurately identify the informativeness of objects in remote-sensing images. 
Talisman performs well on the DOTA-v2.0 dataset. 
However, its performance on the DOTA-v1.0 dataset is relatively poor, and there is also a significant performance gap between Talisman and our method in the early rounds of the DOTA-v2.0 dataset. 
On the other hand, it is worth noting that the BiB method did not perform well on either dataset, especially in the early rounds, where its performance was even worse than that of Random. This may be caused by the fact that BiB was designed for weakly supervised object detection tasks and could not effectively leverage fully supervised knowledge. 
%Our experiments demonstrate that existing active learning methods may not be effective for aerial object detection tasks, highlighting the necessity of designing active learning methods specifically tailored to fully supervised remote sensing object detection tasks.

\begin{table*}[t]
 \begin{center}
  \caption{{Comparison of AP (\%) for ReDet~\cite{han2021redet}} using different AL methods on DOTA-v2.0 \cite{ding2021object} validation set after 5 cycles of active learning. The optimal and second optimal results in each column are highlighted in bold, and the second optimal results are underlined to distinguish them. * denotes that the AL method follows the object-based query type.}
  \label{table:cateAP2}
  \setlength{\tabcolsep}{5pt}{
  \begin{tabular}{c|ccc|cccccc|cccccccc}
   \hline\noalign{\smallskip}
     & \multicolumn{3}{c|}{Common} & \multicolumn{6}{c|}{Middle}& \multicolumn{8}{c}{Rare} \\
   \noalign{\smallskip}
   \hline
   \noalign{\smallskip}
   category                                             & SH                      & SV                    & LV                     & PL                      & HA                      & ST                     & TC                        & BR                     & SP                       & HC                     & BC                     & BD                       & RA                     & SBF                    & GTF                     & CC                        & AP \\
   \noalign{\smallskip}
   \hline
   \noalign{\smallskip}
   Random                                               & 59.2                    & 29.5                  & 37.5                   & 72.5                    & 11.7                    & 47.5                   & 74.4                      & 2.4                    & 23.5                     & 0.0                    & 5.0                    & 18.3                     & 35.3                   & 8.3                    & 19.2                    & 0.0                       & 2.1  \\
   Coreset~\cite{sener2017active}                       & 63.3                    & 29.1                  & 36.3                   & 71.9                    & \bf{17.8}               & 47.9                   & 74.4                      & 3.9                    & 20.4                     & 0.0                    & 2.2                    & 8.5                      & 38.2                   & 9.1                    & 15.5                    & 0.0                       & 0.0 \\
   Localization Stability~\cite{kao2018localization}    & 59.2                    & 28.8                  & 33.3                   & \bf{\underline{76.9}}   & 13.2                    & 48.8                   & 75.8                      & \bf{20.0}              & 21.2                     & 4.3                    & 16.2                   & 33.6                     & 41.5                   & 13.3                   & 29.7                    & 0.0                       & 7.5\\
   Entropy~\cite{settles2009active}                     & 58.7                    & 29.6                  & 37.3                   & 75.6                    & 8.9                     & 47.9                   & 78.6                      & 6.2                    & 20.8                     & \bf{24.0}              & 28.1                   & 36.9                     & 37.5                   & 12.7                   & 29.9                    & 0.0                       & 2.6 \\
   Qbox$^{*}$~\cite{tang2021qbox}                       & \bf{64.6}               & \bf{31.7}             & \bf{44.3}              & 70.9                    & 9.1                     & 47.9                   & 77.9                      & 3.4                    & 22.3                     & 0.0                    & 14.0                   & 17.2                     & 37.9                   & 11.8                   & 32.2                    & 0.0                       & 0.1\\
   BiB~\cite{vo2022active}                              & 58.3                    & 29.8                  & 36.2                   & 75.2                    & 7.5                     & 47.5                   & 79.9                      & 2.3                    & 18.5                     & 0.0                    & 11.1                   & 24.2                     & 35.1                   & 8.9                    & 18.2                    & 0.0                       & 0.0\\
   TALISMAN-FLMI~\cite{kothawade2022talisman}           & 59.9                    &\bf{\underline{30.8}}  &36.3                    & 76.7                    &14.7                     & 48.0                   & 79.2                      &7.5                     & 23.9                     & 0.1                    & 29.1                   & 28.7                     & 41.2                   & 26.4                   & 31.5                    &0.0                        &3.8\\
   TALISMAN-GCMI~\cite{kothawade2022talisman}           & 58.4                    &29.3                   & 33.9                   & \bf{77.7}               & 10.2                    &\bf{52.3}               & 74.9                      & \bf{\underline{19.5}}  & 22.2                     & 5.3                    & 16.8                   & \bf{47.6}                &\bf{49.3}               & 24.6                   & 44.4                    & 0.0                       & \bf{25.7} \\
   \noalign{\smallskip}
   \hline
   \noalign{\smallskip}
   \bf{MUS$^{*}$}                                       & 63.3                    & 30.6                  & 41.3                   & 73.8                    & 15.2                    & 48.6                   & 79.2                      & 7.4                    & \bf{25.8}                & 3.6                    & 37.1                   & 42.4                     & \bf{\underline{45.3}}  & \bf{34.5}              & \bf{\underline{50.2}}   & \bf{0.3}                  & 10.3 \\
   \bf{CDB$^{*}$}                                       & 63.0                    & 30.2                  & \bf{\underline{41.4}}  & 75.3                    & \bf{\underline{17.7}}   & \bf{\underline{49.1}}  & \bf{81.4}                 & 9.6                    & \bf{25.8}                & \bf{\underline{8.1}}   & \bf{42.2}              & \bf{\underline{45.2}}    & 42.5                   & 31.2                   & 49.3                    & 0.0                       & 6.9 \\
   \bf{MUS-CDB$^{*}$}                                   & \bf{\underline{64.0}}   & \bf{\underline{30.8}} & 40.9                   & 74.9                    & 16.3                    & 49.0                   & \bf{\underline{81.2}}     & 11.9                   & \bf{\underline{25.5}}     & 6.9                   & \bf{\underline{41.6}}  & 43.9                     & 45.2                   & \bf{\underline{32.8}}  & \bf{52.5}               & \bf{\underline{0.04}}     & \bf{\underline{13.5}} \\
   \noalign{\smallskip}
   \hline
  \end{tabular}}
 \end{center}
\end{table*}

\begin{table*}[h]
 \begin{center}
  \caption{{Comparison of AP (\%) for ReDet~\cite{han2021redet}} using different AL methods on DOTA-v1.0 \cite{ding2021object} test set after 5 cycles of active learning. 
  The optimal and second optimal results in each column are highlighted in bold, and the second optimal results are underlined to distinguish them.
  * denotes that the AL method follows the object-based query type.}
  \label{table:cateAP1}
  \setlength{\tabcolsep}{5pt}{
  \begin{tabular}{c|ccc|cccccc|cccccc}
   \hline\noalign{\smallskip}
     & \multicolumn{3}{c|}{Common} & \multicolumn{6}{c|}{Middle} & \multicolumn{6}{c}{Rare} \\
   \noalign{\smallskip}
   \hline
   \noalign{\smallskip}
   category                                             & SH                    & SV                    & LV                    & PL                        & HA                    & ST                    & TC                        & BR                        & SP                    & HC                     & BC                    & BD                    & RA                    & SBF                   & GTF                    \\
   \noalign{\smallskip}
   \hline
   \noalign{\smallskip}
   
   Random                                               & 78.9                  & \bf{72.0}             & 68.4                  & 88.3                      & \bf{\underline{54.6}} & 76.8                  & \bf{\underline{90.8}}     & 35.0                      & 56.0                  & 10.6                   & 68.9                  & 58.3                  & 54.6                  & 38.0                  & 49.9                  \\
   Coreset~\cite{sener2017active}                       & \bf{81.0}             & 71.5                  & 66.1                  & 88.1                      & 54.2                  & 75.3                  & \bf{\underline{90.8}}     & 31.8                      & 51.2                  & 24.6                   & 65.6                  & 52.5                  & 51.1                  & 33.7                  & 47.5                  \\
   Localization Stability~\cite{kao2018localization}    & 77.9                  & \bf{72.0}             & 68.0                  & \bf{\underline{88.8}}     & 48.4                  & 76.3                  & \bf{\underline{90.8}}     & \bf{42.4}                 & 54.1                  & 23.0                   & \bf{\underline{73.9}} & 66.6                  & 55.3                  & 41.6                  & 59.8                  \\
   Entropy~\cite{settles2009active}                     & 77.7                  & 70.7                  & 67.2                  & 88.6                      & 45.6                  & 74.5                  & \bf{90.9}                 & 32.5                      & 53.4                  & \bf{37.6}              & \bf{76.7}             & \bf{71.5}             & 54.9                  & 50.5                  & \bf{\underline{62.9}} \\
   Qbox$^{*}$~\cite{tang2021qbox}                       & 78.1                  & \bf{72.0}             & 69.5                  & 87.7                      & 46.6                  & 74.6                  & 90.7                      & 32.3                      & 52.9                  & 23.3                   & 67.6                  & 63.2                  & 50.7                  & 42.8                  & 58.1                  \\
   BiB~\cite{vo2022active}                              & 78.0                  & 71.1                  & 65.2                  & 88.4                      & \bf{54.9}             & 77.5                  & \bf{\underline{90.8}}     & \bf{\underline{39.2}}     & 53.7                  & 22.5                   & 72.6                  & 69.7                  &54.0                   & 46.7                  & 54.9                  \\
   TALISMAN-FLMI~\cite{kothawade2022talisman}           & 78.0                  & 71.5                  & \bf{67.1}             & \bf{88.9}                 & 47.6                  & \bf{\underline{78.6}} & 90.7                      & 35.5                      & 51.9                  & 15.0                   & 68.2                  & 58.7                  & 52.3                  & 35.1                  & 50.4                  \\
   TALISMAN-GCMI~\cite{kothawade2022talisman}           & 78.2                  & 71.2                  & 66.0                  & \bf{\underline{88.8}}     & 50.4                  & \bf{78.7}             & 90.7                      & 35.6                      & 55.5                  & 21.3                   & 69.1                  & 61.7                  & 52.6                  & 37.7                  & 54.7                  \\
   
    \noalign{\smallskip}
   \hline
   \noalign{\smallskip}

   \bf{MUS$^{*}$}                                       & 81.1                  & \bf{\underline{71.7}} & \bf{\underline{69.6}} & 88.3                      & 53.1                  & 77.4                  & 90.7                      & 37.2                      & 54.9                  & 32.8                   & 72.4                  & 68.9                  & 55.7                  & \bf{\underline{50.9}} & 62.1                  \\
   \bf{CDB$^{*}$}                                       & \bf{\underline{79.5}} & 71.6                  & 69.2                  & 88.5                      & 54.0                  & 78.2                  & 90.7                      & 36.8                      & \bf{59.1}             & \bf{\underline{33.2}}  & 72.1                  & 69.4                  & \bf{57.4}             & 50.7                  & 62.6                  \\
   \bf{MUS-CDB$^{*}$}                                   & \bf{\underline{79.5}} & \bf{\underline{71.7}} & 69.2                  & \bf{\underline{88.8}}     & 53.3                  & 78.5                  & 90.7                      & 38.8                      & \bf{\underline{57.3}} & 33.1                   & \bf{\underline{73.9}} & \bf{\underline{70.8}} & \bf{55.9}             & \bf{51.2}             & \bf{63.4}             \\
   \noalign{\smallskip}
   \hline
  \end{tabular}}
 \end{center}
\end{table*}

For the detectors KLD~\cite{yang2021learning} and SASM~\cite{hou2022shape}, the experiments were conducted on DOTA-v2.0~\cite{ding2021object} validation set, and we compared the proposed method with four other commonly used active learning methods, including random, Coreset, Localization Stability, and entropy. 
We did not compare our method with the Qbox, BiB, and Talisman methods on the KLD~\cite{yang2021learning} and SASM~\cite{hou2022shape} detectors. 
This is because single-stage detectors do not output background scores and prediction box features necessary for calculating sample information using the Qbox, BiB, and Talisman methods. 
%Therefore, these methods cannot be used on single-stage detectors. 
The experimental results are presented in Table~\ref{tab:result}. 
The proposed method significantly improved both KLD~\cite{yang2021learning} and SASM~\cite{hou2022shape} detectors, demonstrating the generality and effectiveness of the proposed approach on different types of detectors. 

Regarding the two-stage detector Oriented R-CNN~\cite{xie2021oriented}, we evaluated its performance using different AL methods on the DOTA-v2.0 validation set, and the mAP values are presented in Table~\ref{tab:result}. 
%We can see that our method outperforms other active learning methods in terms of mAP while sampling the same number of objects.
%The performance improvement is significant, demonstrating the effectiveness and generality of our approach across different types of detectors. 
%We also observed that representativeness-based AL methods, such as random and Coreset, do not perform well for remote sensing scenes. 
%On the other hand, uncertainty-based AL methods are more effective for these scenes. The poor performance of Localization Stability and Qbox in remote sensing scenes suggests the need for designing appropriate AL methods for this particular scenario. 

In summary, our experiments on multiple detectors demonstrate the effectiveness and generalizability of the proposed method MUS-CDB. 
The proposed approach can be easily integrated into various object detection frameworks and can help improve the performance of object detection models in various applications.}

\subsubsection{Category-wise AP} 
Table \ref{table:cateAP2} and \ref{table:cateAP1} show the average category-wise AP of all the compared methods using ReDet detector, on DOTA-v1.0 and DOTA-v2.0, respectively. 
%The optimal and second-best results in each column are highlighted in boldface, while the second-best result is underlined.
From Table \ref{table:cateAP2} and \ref{table:cateAP1}, we can observe that MUS-CDB and its variants can usually achieve the best performance in most categories. 
%Especially some categories that are not well-learned. This result indicates that our proposed query strategy can impartially improve the performance of each category. 

{Qbox and uncertainty-based active learning methods tend to excessively query specific categories while ignoring the overall performance improvement. 
For example, Qbox focuses on improving the performance of [bridge], [tennis court], and [storage tank] but neglects the performance improvement of rare categories. 
Similarly, entropy-based methods only improve performance for a few middle and rare categories, such as [helicopter] and [container crane], while paying little attention to other categories, such as [plane], [swimming pool], and [ground track field]. Furthermore, it is worth noting that BiB could not accurately query rare categories for annotation, while TALISMAN-GCMI could only find and annotate a limited number of rare categories. 
In contrast, our method can effectively improve the performance of all categories, including rare categories. This highlights the effectiveness and versatility of our proposed method for remote sensing object detection tasks.}

\begin{figure*}[!t]
\centering
\includegraphics[width=\textwidth]{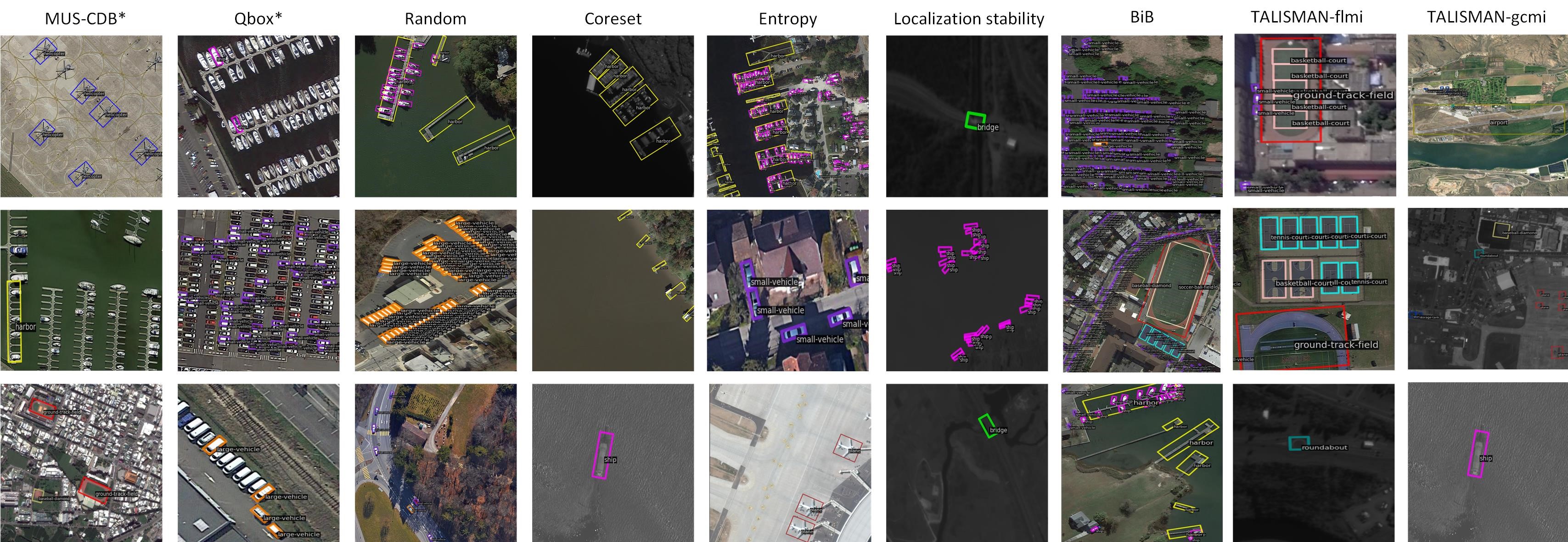}
\caption{Examples of the top-rated objects or images queried by MUS-CDB, Qbox \cite{tang2021qbox}, Random, Coreset \cite{sener2017active}, Entropy \cite{settles2009active}, Localization Stability \cite{kao2018localization}, BiB~\cite{vo2022active}, TALISMAN-flmi~\cite{kothawade2022talisman} and TALISMAN-gcmi~\cite{kothawade2022talisman} on the DOTA-v2.0~\cite{ding2021object}. 
* denotes that the AL method follows the object-based query type. 
The visual comparison results in the first and second columns show that our proposed method MUS-CDB can sample more valuable middle and rare classes than another object-based sampling method Qbox. 
In addition, compared with other image-based sampling methods, it can be found that MUS-CDB can effectively save sampling costs through object-based sampling and then sampling more diverse objects.}
\label{vis}
\end{figure*}

\begin{table}[h]
\centering
\caption{Comparison of mAP(\%) for ReDet~\cite{han2021redet} on DOTA-v2.0 ~\cite{ding2021object} validation set with different score thresholds in Equ.~\ref{eq:image} of Mixed Uncertainty Sampling module (MUS). {The Cycle-0 column reports the performance without any active learning sampling strategy.}}
\begin{tabular}{c|ccccc} 
\hline\noalign{\smallskip}
$\theta$& Cycle-0 & Cycle-1 & Cycle-2 & Cycle-3 & Cycle-4 \\
\noalign{\smallskip}
\hline
\noalign{\smallskip}
0.05 & 22.8 & 30.8 & 36.7 & 39.4 & 40.6 \\
%0.80 & 22.8 & 32.9 & \bf{38.3} & \bf{40.9} & \bf{41.2} \\
0.10 & 22.8 & \bf{34.1} & \bf{37.2} & \bf{40.2} & \bf{40.9} \\
0.15 & 22.8 & 32.7 & 37.1 & 39.4 & 40.4 \\
\noalign{\smallskip}
\hline
\end{tabular}
\label{table:threshold}
\end{table}

\subsubsection{Score Threshold $\theta$} 
$\theta$ is the parameter of module MUS in Equ.~\eqref{eq:image}, which is used to impose preference on the images with distinct information in the active selection phase. 
According to the results in Table \ref{table:threshold}, the optimal value of $\theta$ is obtained around 0.10.
Therefore, we use $\theta$ = 0.10 for all our experiments.
This can be explained by the fact that when $\theta$ is close to the lower bound of 0.05, more noise is involved in calculating the image uncertainty score. 
Therefore the final object information calculation is inaccurate. 
Conversely, if $\theta$ becomes larger, the performance degrades. 
This is because predicted bounding boxes containing rare patterns in the image are filtered because of low predicted scores and cannot participate in calculating image information.

\subsubsection{Visualization of the Queried Boxes}
To demonstrate whether MUS-CDB selects the expected objects for sampling, we visualize the top-ranked objects sampled by the object-based sampling method MUS-CDB versus Qbox and the top-ranked images sampled by the image-based sampling methods random, entropy, Coreset, and localization stability. 
%The visualization results are listed in Figure~\ref{vis}.
%There is a long tail phenomenon in the DOTA-v1.0 and DOTA-v2.0 datasets, where we classify the object categories into common, middle, and rare categories based on their distribution. 
%The detailed classification of object categories is shown in Table~\ref{table:cateAP2} and Table~\ref{table:cateAP1}. 
According to Figure~\ref{vis}, we can see that our method MUS-CDB can preferentially sample informative examples of rare categories, such as [helicopter], [baseball diamond], [ground track field], and [soccer-ball field]. 
The sampling results of the random method are consistent with the original category distribution of the DOTA-v2.0 dataset, manifested in over-sampling the low-informative head categories and intermediate categories such as [plane], [ship], [small vehicle], and [large vehicle]. 
%Although the collected images have rich land types, such as airports, ports, and cities, the distribution of the queried samples is too similar to the labeled dataset, which does not improve the model performance very much. 
As an object-based sampling method, Qbox oversamples common categories, such as the classes [small vehicle], [large vehicle], and [ship], and fails to sample rare categories. 
%Although informative samples in the common category are collected, these samples do not improve the model performance much. 
Entropy (the image-based sampling method) results in most of the queried samples being in the same pattern, leading to information redundancy. 
The images sampled by the Coreset and Localization Stability methods have a single land type. The Coreset method preferentially samples images in the ocean background, and the Localization Stability method preferentially samples images in the dark scenes. The sample distribution in the images collected by the two methods is relatively sparse, which can effectively alleviate the problem of redundant annotation. 
However, the two methods pay too little attention to the rare category, resulting in a low-performance improvement of the final model. 
The BiB method is prone to over-sampling densely distributed images, which is not a significant issue in generic scenarios where objects are sparsely distributed in images. 
However, in remote sensing scenarios, this can lead to redundant sampling due to the dense distribution of the same object categories in the images, which can limit the effectiveness of the sampling method in improving the overall performance of the final model. 
The TALISMAN method introduces rare categories as reference objects to help the sampling method collect more samples of rare categories. However, compared to our proposed method, it lacks an overall improvement in performance for all categories.
%These results demonstrate that our proposed method can achieve cost-effective querying in remote sensing scenarios.

%--------------------------
\begin{table}[h]
\centering
\caption{{Ablation studies on strategies of Mixed Uncertainty Sampling module (MUS), Class Distribution Balancing module (CDB) based on mAP(\%). The Cycle-0 column reports the performance without any active learning sampling strategy.}}
\setlength{\tabcolsep}{1mm}{
\begin{tabular}{ccccccccc} 
\hline 
\noalign{\smallskip}
{Method}&MUS&CDB& Cycle-0 & Cycle-1 & Cycle-2 & Cycle-3 & Cycle-4\\
\noalign{\smallskip}
\hline
\noalign{\smallskip}
{Random}&&&22.8&24.6&25.1&25.3&26.2 \\
\hline
\noalign{\smallskip}
(a)&\checkmark&&22.8&33.0&36.2&38.1&39.9\\
(b)&&\checkmark&22.8&32.9&37.0&37.3&40.5 \\
% (c)&\checkmark&\checkmark&22.8&32.4&36.3&38.3&39.9 \\
(d)&\checkmark&\checkmark&22.8&\bf{34.1}&\textbf{37.2}&\textbf{40.2}&\textbf{40.9}\\
\noalign{\smallskip}
\hline
\end{tabular}}
\label{table:ablation}
\end{table}

\subsection{Ablation Study}

\subsubsection{Evaluation on each component of MUS-CDB}
To further investigate the effectiveness of each component of MUS-CDB, we conduct ablation experiments on the DOTA-v2.0 task. As shown in Table \ref{table:ablation}, satisfactory and consistent gains from the Random baseline to our whole method demonstrate the validity of each module. From Table \ref{table:ablation} we can see that both MUS and CDB module has achieved significant performance improvements compared to the Random baseline. {The results in Table \ref{table:ablation} demonstrate that both the MUS and CDB modules contribute significantly to the overall performance of MUS-CDB, and together they provide a more effective and efficient way of selecting highly informative predicted bounding boxes for training object detection models.}

\begin{table}[h]
\centering
\caption{{Comparison of mAP(\%) for ReDet~\cite{han2021redet}  on DOTA-v2.0~\cite{ding2021object} validation set for evaluating the effectiveness of MUS Method via ablation experiment: Entropy-Based Uncertainty Sampling (Method 1) vs. MUS Method (Method 2). The Cycle-0 column reports the performance without any active learning sampling strategy.}}
\begin{tabular}{c|ccccc} 
\hline\noalign{\smallskip}
Method& Cycle-0 & Cycle-1 & Cycle-2 & Cycle-3 & Cycle-4 \\
\noalign{\smallskip}
\hline
\noalign{\smallskip}
(1) &22.8&30.8&35.2&36.2&38.4 \\
(2) &22.8&\bf{33.0}&\textbf{36.2}&\textbf{38.1}&\textbf{39.9} \\
\noalign{\smallskip}
\hline
\end{tabular}
\label{table:image-guided}
\end{table}

\subsubsection{Evaluation on the image-level term of MUS}
{To further investigate the effectiveness of the proposed method, we conducted ablation experiments specifically targeting the image-level term proposed in the MUS. We conducted these experiments on the DOTA-v2.0 dataset, and the results are summarized in Table \ref{table:image-guided}. As we can see from the table, incorporating global information into the calculation of sample informativeness (i.e., adding the image-level term to the original entropy-based uncertainty measurement) leads to a significant improvement in detection performance compared to using the original entropy-based uncertainty measurement alone. This improvement demonstrates that incorporating spatial and semantic information from the image can benefit object detection tasks in remote sensing scenarios. The proposed MUS module is a valuable addition to the MUS-CDB framework. In summary, our ablation experiments provide strong evidence supporting the effectiveness of the proposed image-level term in the MUS module and demonstrate the importance of considering both the image-level and instance-level information in the MUS-CDB framework for achieving better object detection performance in remote sensing applications.}

\begin{figure}[htbp]
\centering
\includegraphics[width=0.5\textwidth]{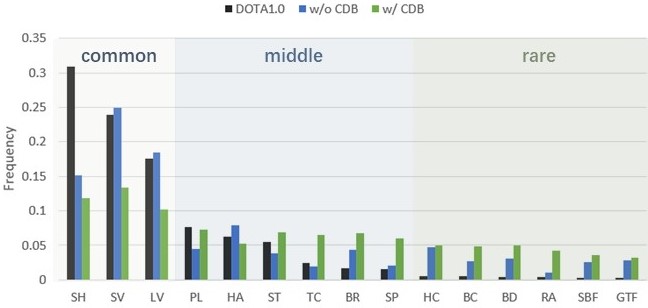}
\caption{The average category frequency (\%) of DOTA-v1.0 \cite{xia2018dota} sampled by our AL methods in each round with ReDet~\cite{han2021redet}. Our proposed Class Distribution Balancing module (CDB) can sample more middle and rare and fewer common categories.}
\label{class}
\end{figure}

\begin{figure}[h]
\centering
\includegraphics[width=0.5\textwidth]{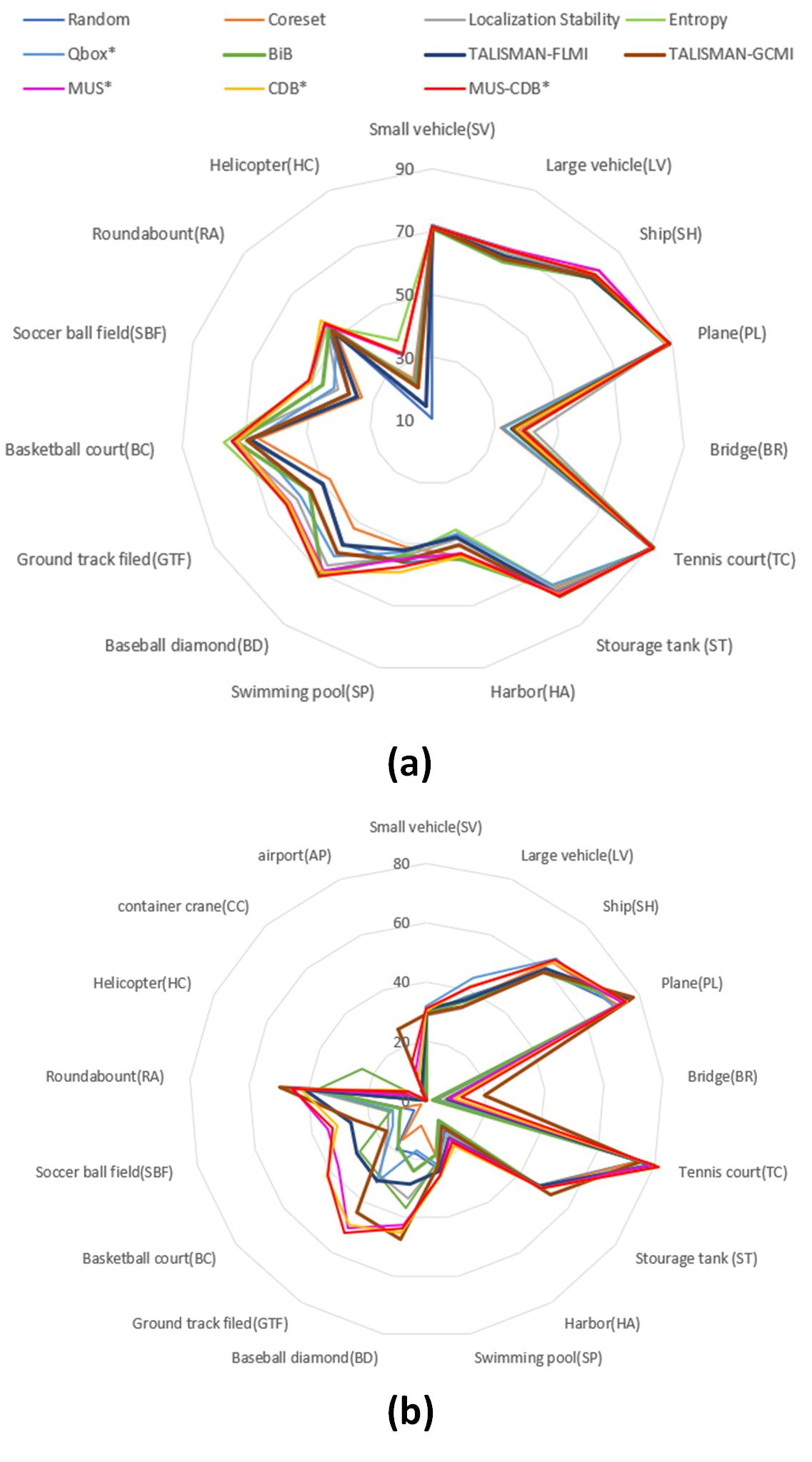}
\caption{Radar charts for each category of objects in different datasets. 
(a) DOTA-v1.0 \cite{xia2018dota}. 
(b) DOTA-v2.0 \cite{ding2021object}. 
Different colors represent different active learning methods. 
The larger area enclosed by the outer line, the better performance of the corresponding method. 
The value in this graph denotes the average AP over five cycles of active learning.}
\label{radar}
\end{figure}

\subsubsection{Analysis of Queried Class Distribution using CDB Method} 
Figure~\ref{class} shows the class frequency of the DOTA-v1.0 dataset and the average class frequency of training samples selected by our methods in each cycle of AL. 
The actual distribution of each class in the DOTA-v1.0 is exactly long-tailed. 
Most AL methods tend to select many common categories for annotation, resulting in ineffective and redundant sampling. 
The CDB module in our method enables us to allocate this problem in middle and rare categories such as [tennis court], [bridge], [swimming pool], [basketball court], [baseball diamond], and [roundabout] as shown in the green column of Figure~\ref{class}. 
This approach produces more diverse and balanced samples for annotation and avoids being influenced by the dataset's distribution. This improvement demonstrates the unique contribution of the CDB module to our method, which other active learning methods may not have considered. Our method provides a more effective and efficient approach for addressing the long-tail distribution problem in object detection datasets. The performance superiority of our method over other methods in each category is visualized in Figure~\ref{radar}.

%--------------------------

\begin{table}[t]
\centering
\caption{{Comparison of mAP(\%) for ReDet~\cite{han2021redet} on DOTA-v2.0~\cite{ding2021object} validation set for the MUS-CDB Method using ReDet's Loss Function and Improved Loss Function: Method 1 with Default Loss Function vs. Method 2 with Proposed Improved Loss Function. The Cycle-0 column reports the performance without any active learning sampling strategy.}}
\begin{tabular}{c|ccccc} 
\hline\noalign{\smallskip}
Method& Cycle-0 & Cycle-1 & Cycle-2 & Cycle-3 & Cycle-4 \\
\noalign{\smallskip}
\hline
\noalign{\smallskip}
(1) &22.8&32.4&36.3&38.3&39.9 \\
(2) &22.8&\bf{34.1}&\textbf{37.2}&\textbf{40.2}&\textbf{40.9} \\
\noalign{\smallskip}
\hline
\end{tabular}
\label{table:loss}
\end{table}

\subsubsection{Evaluation on proposed loss function}
{To further evaluate the proposed loss function's effectiveness in the MUS-CDB context, we conducted experiments on DOTA-v2.0 to compare the performance of MUS-CDB using the original loss function and the proposed modified loss function. The experimental results are presented in Table \ref{table:loss}. As we can see from the table, using the proposed modified loss function leads to a significant improvement in detection performance compared to using the original loss function. This improvement demonstrates the effectiveness of the proposed loss function in reducing the noise introduced by partially annotated images during model training and improving the robustness of the model to such noisy inputs. One possible reason why the proposed loss function is effective is that it adaptively adjusts the classification loss weight for background samples. This helps the model assign less weight to the background samples likely to be noisy or ambiguous and more weight to the informative samples that can contribute to the model training more effectively. This can help reduce the impact of noisy and ambiguous samples on the model training and improve the overall performance of the model.}

%-------------------------
\section{Conclusion}\label{Conclusion}
In this paper, we propose an object-based active learning method MUS-CDB to alleviate the massive burden of aerial object detection data annotation. We devise an image- and object-guided uncertainty sampling selection criterion in active querying to identify the most informative instances. We consider the long-tailed problem of the remote sensing image dataset and impose class preference during active sampling to promote the diversity of selected objects. An effective training method for partially labeled data is also proposed to utilize the queried knowledge. Extensive experiments on the DOTA-v1.0 and DOTA-v2.0 benchmarks demonstrate the superiority of the proposed MUS-CDB. 
%In the future, we will further study the effectiveness of considering the angle information in the active selection for aerial object detection.
\section*{Acknowledgment}

The authors thank Professor Gui-Song Xia and Dr. Jiang Ding from Wuhan University for their helpful discussion and for solving the problem using DOTA 2.0 Dataset. This work is partly supported by the National Natural Science Foundation of China under grant 62272229, and the Natural Science Foundation of Jiangsu Province under grant BK20222012. 
Dong Liang, Jing-Wei Zhang, Ying-Peng Tang  are contributed equally to this work. Corresponding author: Sheng-Jun Huang (huangsj@nuaa.edu.cn). 
%\section{References Section}
\bibliographystyle{ieeetr}  % ieeetr国际电气电子工程师协会期刊
\bibliography{reference}  % ref就是之前建立的ref.bib文件的前缀

%\section*{Biography Section}
\begin{IEEEbiography}[{\includegraphics[width=1in,height=1.25in,clip,keepaspectratio]
{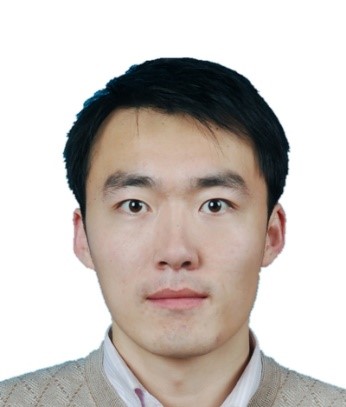}}]{Dong Liang}
received a B.S. degree in Telecommunication Engineering and an M.S. in Circuits and Systems from Lanzhou University, China, in 2008 and 2011, respectively. In 2015, he received Ph.D. at the Graduate School of IST, Hokkaido University, Japan. He is an associate professor at the College of Computer Science and Technology, Nanjing University of Aeronautics and Astronautics. His research interests include pattern recognition and image processing. He was awarded the Excellence Research Award from Hokkaido University in 2013. He has published several research papers including in IEEE TIP/TNNLS/TMM/TGRS/TCSVT, Pattern Recognition, AAAI, and IJCAI.
%\vspace{-4mm}
\end{IEEEbiography}
%-------------------------------------------------------------------------
\begin{IEEEbiography}[{\includegraphics[width=1in,height=1.25in,clip,keepaspectratio]{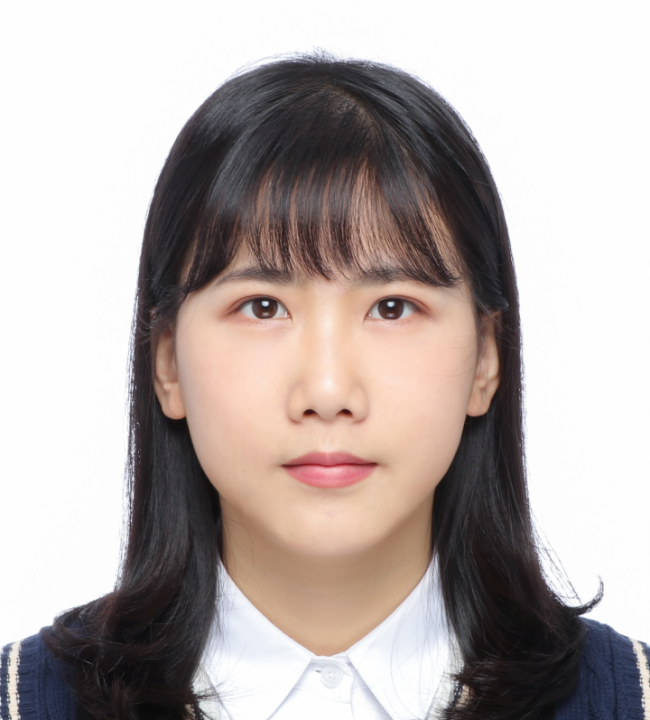}}]{Jing-Wei Zhang}
received the B.S. degree in computer science and technology from Jiangsu University, Zhenjiang, China, in 2021. She is pursuing a master’s degree with the College of Computer Science and Technology, Nanjing University of Aeronautics and Astronautics, Nanjing. Her research interests include active learning and object detection.
\end{IEEEbiography}
%-------------------------------------------------------------------------
\begin{IEEEbiography}[{\includegraphics[width=1in,height=1.25in,clip,keepaspectratio]{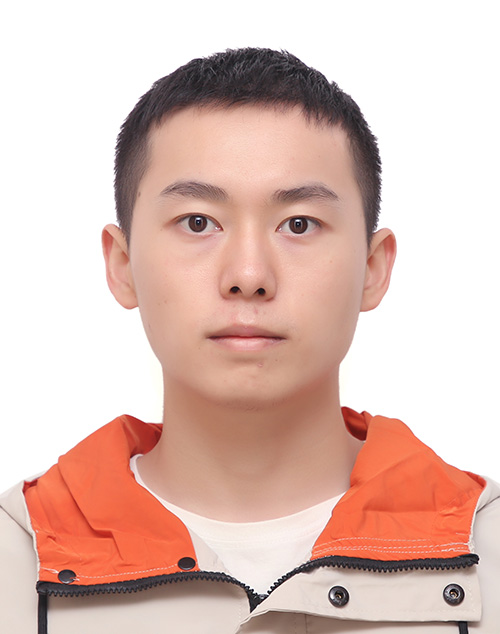}}]
{Ying-Peng Tang} 
	received the BSc degree from the Nanjing University of Aeronautics and Astronautics, China, in 2020. He is currently pursuing a Ph.D. degree in computer science and technology with the Nanjing University of Aeronautics and Astronautics, Nanjing, China. His current research interests include active learning and semi-supervised learning. He was awarded for China National Scholarship in 2022 and the Excellent Master thesis in Jiangsu Province in 2021.
\end{IEEEbiography}
\begin{IEEEbiography}[{\includegraphics[width=1in,height=1.25in,clip,keepaspectratio]{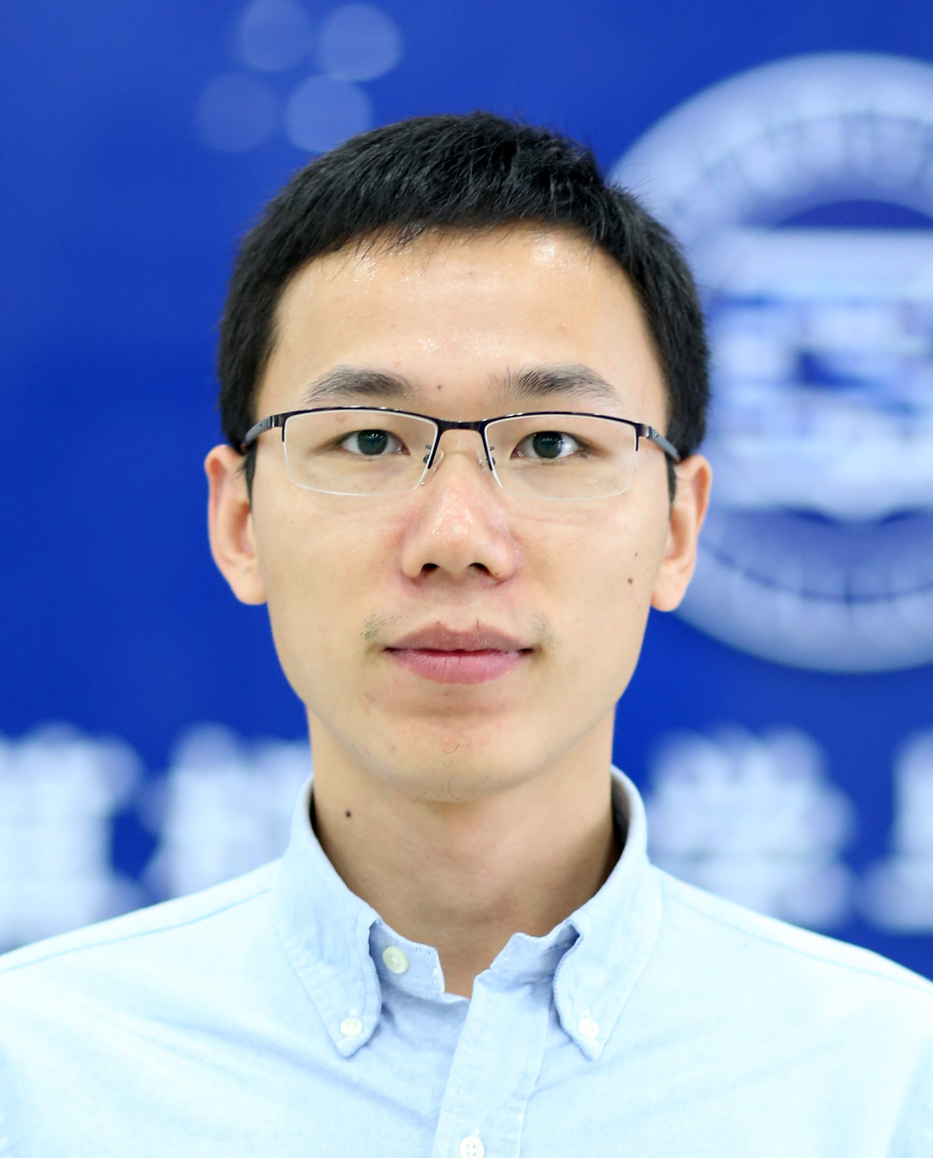}}]{Sheng-Jun Huang}
received the BSc and Ph.D.
degrees in computer science from Nanjing University, China, in 2008 and 2014, respectively.
He is now a professor at the College of Computer
Science and Technology at Nanjing University
of Aeronautics and Astronautics. His main research interests include machine learning and
data mining. He has been selected for the Young
Elite Scientists Sponsorship Program by CAST
in 2016, and won the China Computer Federation Outstanding Doctoral Dissertation Award in
2015, the KDD Best Poster Award in 2012, and the Microsoft
Fellowship Award in 2011. He is a Junior Associate Editor of Frontiers
of Computer Science.
\end{IEEEbiography}
%\vspace{11pt}
%\vfill

\end{document}